\documentclass[twoside,11pt]{article}
\usepackage{jmlr2e}
\usepackage{amsmath,verbatim,amsfonts,amscd}
\usepackage{latexsym,graphicx, bm, float, color}
\usepackage{graphics,mathtools,hyperref}
\usepackage{dsfont}
\usepackage{framed}
\usepackage[most]{tcolorbox}
\usepackage{xcolor}
\usepackage{enumitem}

\newcommand{\N}{\mathcal{N}}

\newcommand{\x}{\textbf{x}}
\newcommand{\y}{\textbf{y}}
\newcommand{\z}{\textbf{z}}
\newcommand{\m}{\mathfrak{m}}

\firstpageno{1}

\begin{document}

\title{Trainability of ReLU networks and Data-dependent Initialization}

\author{\name Yeonjong Shin \email yeonjong\_shin@brown.edu \\
       \name George Em Karniadakis \email george\_karniadakis@brown.edu \\
       \addr Division of Applied Mathematics\\
       Brown University\\
       Providence, RI 02912, USA}

\editor{}

\maketitle

\begin{abstract}
	In this paper, we study the trainability of rectified linear unit (ReLU) networks.
	A ReLU neuron is said to be dead if it only outputs a constant for any input.
	Two death states of neurons are introduced; tentative and permanent death.
	A network is then said to be trainable if the number of permanently dead neurons is sufficiently small for a learning task.
	We refer to the probability of a network being trainable as trainability.
	We show that a network being trainable is a necessary condition for  successful training and 
	the trainability serves as an upper bound of successful training rates.
	In order to quantify the trainability, we study the probability distribution of the number of active neurons at the initialization.
	In many applications, over-specified or over-parameterized neural networks are successfully employed and shown to be trained effectively. 
	With the notion of trainability, we show that over-parameterization is both a necessary and a sufficient condition for minimizing the training loss.
	Furthermore, we propose a data-dependent initialization method in the over-parameterized setting. 
	Numerical examples are provided to demonstrate the effectiveness of the method and our theoretical findings.
\end{abstract}


\section{Introduction} 
\label{sec:introduction}
Neural networks have been successfully used in various fields of applications.
These include image classification in computer vision \citep{krizhevsky2012imagenet},
speech recognition \citep{hinton2012deep}, natural language translation \citep{wu2016google}, and superhuman performance in the game of Go \citep{silver2016mastering}.
Modern neural networks are often severely over-parameterized or over-specified. 
Over-parameterization means that the number of parameters is much larger than the number of training data.
Over-specification means that the number of neurons in a network is much larger than needed.
It has been reported that 
the wider the neural networks, 
the easier it is to train
\citep{livni2014computational, safran2016quality, nguyen2017loss}.

In general, neural networks are trained by first- or second-order gradient-based optimization methods from random initialization. 
Almost all gradient-based optimization methods are stemmed from backpropagation \citep{rumelhart1985learning} and the stochastic gradient descent (SGD) method \citep{robbins1951stochastic}. 
Many variants of vanilla SGD have been proposed. For example, AdaGrad \citep{duchi2011adaptive}, RMSProp \citep{hintonlecture6a}, Adam \citep{kingma2014adam}, AMSGrad \citep{reddi2019convergence}, and L-BFGS \citep{byrd1995limited}, to name just a few. 
Different optimization methods have different convergence properties. 
It is still far from clear 
how different optimization methods affect the performance of trained neural networks.
Nonetheless, 
how to start the optimization processes plays a crucial role for the success of training.
Properly chosen weight initialization
could drastically improve the training performance and allow the training of deep neural networks, for example, see
\citep{lecun1998efficient,glorot2010understanding,saxe2013exact, he2015delving, mishkin2015all}, and for more
recent work see \citep{lu2019dying}.
Among them, when it comes to the rectified linear unit (ReLU) neural networks, the `He initialization' \citep{he2015delving} is one of the most commonly used initialization methods.

\begin{figure}[!htbp]
	\centerline{
		\includegraphics[width=6cm]{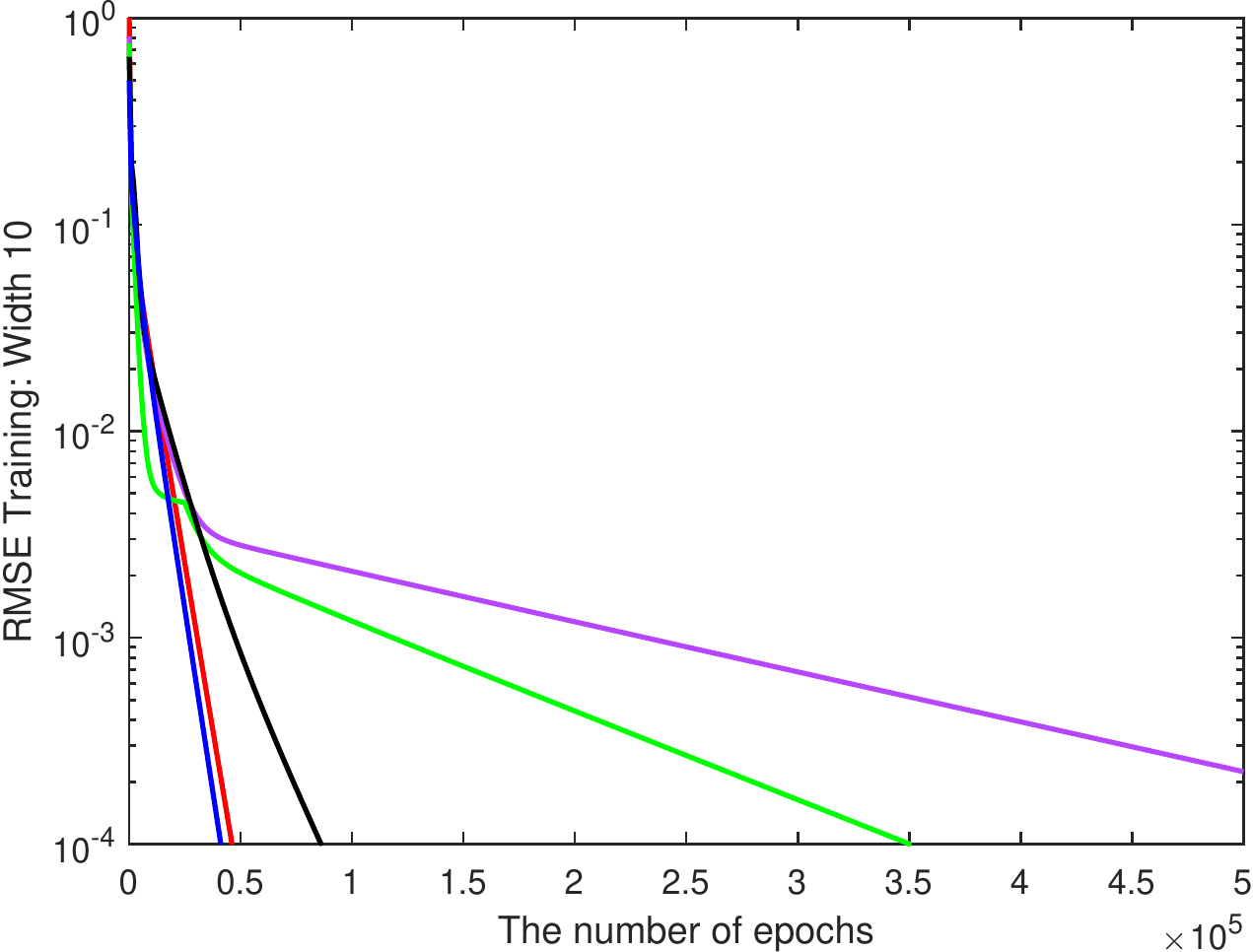}
		\includegraphics[width=6cm]{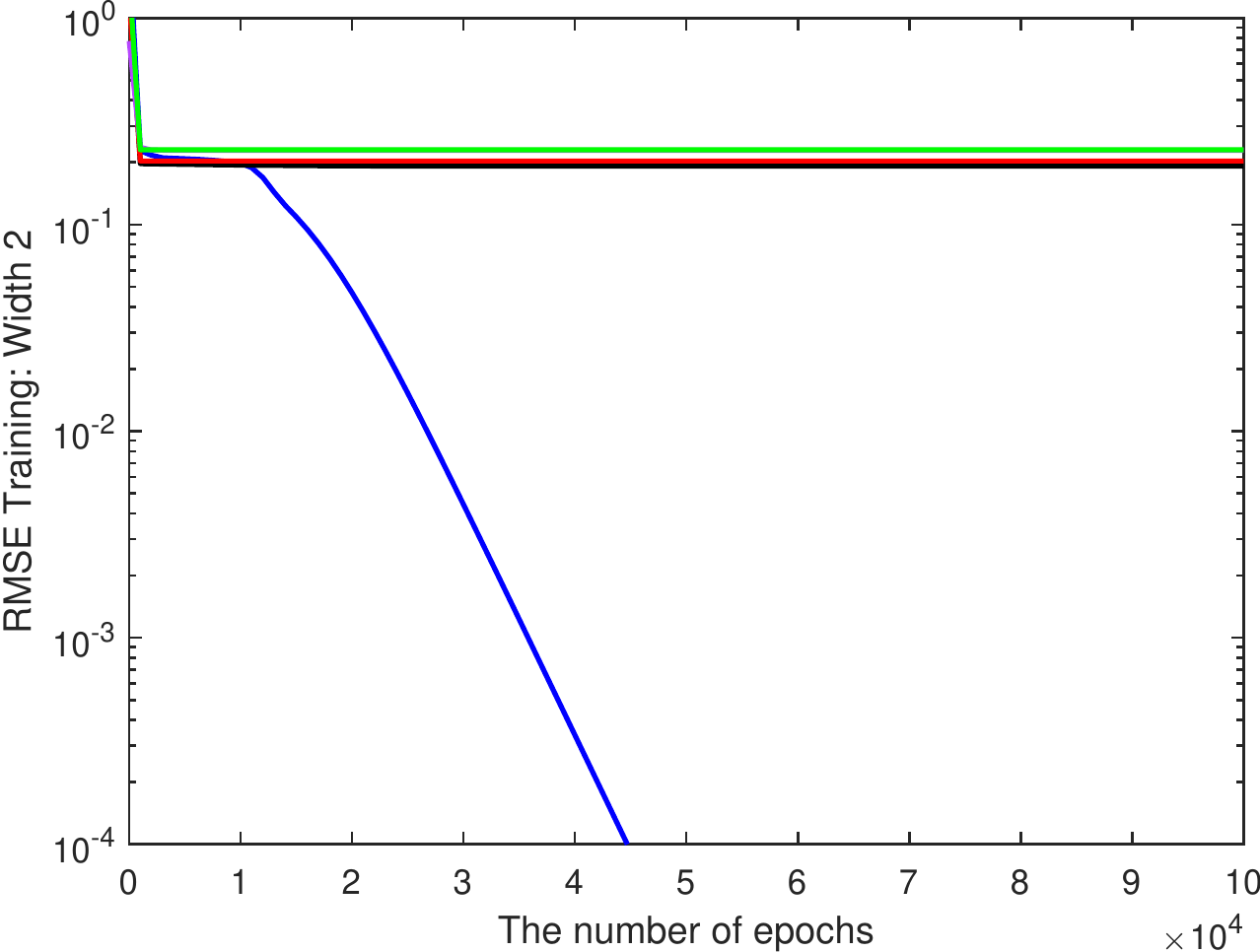}
	}
	\caption{The root mean square errors on the training data of five independent simulations with respect to the number of epochs. The standard $L_2$-loss is employed.
	(Left) Width 10 and depth 2. (Right) Width 2 and depth 2.}
	\label{fig:motivation}
\end{figure}

There are several theoretical works showing that under various assumptions, over-parameterized neural networks can perfectly interpolate the training data. 
For the shallow neural network setting, see
\citep{oymak2019towards,soltanolkotabi2019theoretical, du2018gradient-shallow, li2018learning}.
For the deep neural network setting, see
\citep{du2018gradient-DNN, zou2018stochastic, allen2018convergence}.
Hence, over-parameterization can be viewed as a sufficient condition for minimizing the training loss.
Despite of the current theoretical progress, 
there still exists a huge gap between existing theories and empirical observations 
in terms of the level of over-parameterization.
To illustrate this gap, let us consider the problem of approximating $f(x) = |x|$.
The same learning task was also used in \cite{lu2019dying}, but with a deep network.
Here, we consider a shallow ReLU network.
The training set consists of 10 random samples from the uniform distribution on $[-1,1]$.
To interpolate all 10 data points, the best existing theoretical condition requires the width of $\mathcal{O}(n^2)$ \citep{oymak2019towards}.
In this case, the width of 100 would be needed.
Figure~\ref{fig:motivation} shows the convergence of the root mean square errors (RMSE) on the training data with respect to the number of epochs for five independent simulations.
On the left, the results of width 10 are shown.
We observe that all five training losses converge to zero as the number of epochs increases.
It would be an ongoing challenge to bridge the gap of the degree of over-parameterization.
On the other hand, we know that $f(x)=|x|$ can be exactly represented by only two ReLU neurons as $|x| = \max\{x,0\} + \max\{-x,0\}$.
Thus, we show the results of width 2 on the right of Figure~\ref{fig:motivation}.
In contrast to the theoretical guarantee, we observe that 
only one out of five simulations can achieve zero training error. 
It turns out that there is a probability greater than 0.43 that the network of width 2 fails to be trained successfully (Theorem~\ref{THM:MAIN}); see also \citep{lu2019dying}.


In this paper, we study the trainability of ReLU networks, a {\em necessary} condition for successful training
and propose a data-dependent initialization for the better training.
Our specific contributions are summarized below:
\begin{itemize}
	\item We classify a dead neuron into two states; tentatively dead and permanently dead.
	With the new classification, we introduce a notion of trainable networks (precise definition is given in section~\ref{sec:trainability}). 
	By combining it with Lemma~\ref{THM:DEAD-STAY-DEAD}, we conclude that a network being trainable is a necessary condition for successful training.
	That is, if an initialized ReLU network is not trainable,
	regardless of which gradient-based optimization method is selected, the training will not be successful.
	\item The probability of a network being trainable after random initialization is referred to as {\em trainabilty} (trainable probability).
	We establish a general formulation of computing trainability 
	and derive the trainabilities of ReLU networks of depth $2$ and $3$. 
	\item With the computed trainability, 
	we show that for shallow ReLU networks, over-parameterization is both 
	\textit{a necessary and a sufficient condition} for minimizing the training loss, i.e., interpolating all training data.
	\item Motivated by our theoretical results, we propose a new data-dependent initialization scheme.
\end{itemize}

Taken together, our developments provide new insight into the training of ReLU neural networks 
that can help us design efficient network architectures and reduce the effort in optimizing the networks.

The rest of this paper is organized as follows.
Upon presenting the mathematical setup in section~\ref{sec:setup},
we present the trainability of ReLU networks in section~\ref{sec:trainability}. 
A new data-dependent initialization is introduced in section~\ref{sec:data-dependent}.
Numerical examples are provided in section~\ref{sec:example},
before the conclusion in section~\ref{sec:conclusion}.


\section{Mathematical Setup} \label{sec:setup}

Let $\mathcal{N}^L:\mathbb{R}^{d_{\text{in}}} \mapsto \mathbb{R}^{d_{\text{out}}}$ be a feed-forward neural network with $L$ layers and $n_j$ neurons in the $j$-th layer ($n_0 = d_{\text{in}}=d$, $n_{L} = d_{\text{out}}$).
For $1\le j \le L$, 
the weight matrix and the bias vector in the $j$-th layer are denoted by $\bm{W}^{j} \in \mathbb{R}^{n_j \times n_{j-1}}$ and $\bm{b}^{j} \in \mathbb{R}^{n_j}$, respectively;
$n_j$ is called the width of the $j$-th layer.
We also denote the input by $\textbf{x} \in \mathbb{R}^{d_{\text{in}}}$
and the output at the $j$-th layer by 
$\mathcal{N}^{j}(\textbf{x})$.
Given an activation function $\phi$ which is applied element-wise,
the feed-forward neural network is defined by
\begin{align*}
	\mathcal{N}^{j}(\textbf{x}) &= \bm{W}^{j}\phi(\mathcal{N}^{j-1}(\textbf{x})) + \bm{b}^{j}
	\in \mathbb{R}^{n_j}, 
	\qquad \text{for} \quad 2 \le j \le L,
\end{align*}
and $\mathcal{N}^1(\textbf{x}) = \bm{W}^{1}\textbf{x} + \bm{b}^{1}$.
Note that $\N^L(\x)$ is called a $(L-1)$-hidden layer neural network or a $L$-layer neural network.
Also, $\phi(\N_i^j(\x))$, $i=1,\cdots,n_j$, is called a neuron or a unit in the $j$-th hidden layer. 
We use $\bm{n} = (n_0,\cdots,n_L)$ to describe a network architecture.
In this paper, we refer a 2-layer network as a shallow network
and a $L$-layer network as a deep network for $L > 2$.

Let $\bm{\theta}$ be a collection of all weight matrices and bias vectors, i.e., 
$\bm{\theta} = \{\bm{V}^j\}_{j=1}^L$ where $\bm{V}^j=[\bm{W}^j, \bm{b}^j]$.
To emphasize the dependency on $\bm{\theta}$, we often denote the neural network 
by $\N^L(\x;\bm{\theta})$.
In this paper, the rectified linear unit (ReLU) is employed as an activation function, i.e,
$$
\phi(\x) = \text{ReLU}(\x) :=\left(\max\{x_1, 0\}, \cdots, \max\{x_{d_{\text{in}}}, 0\} \right)^T,
$$
where $\x = (x_1,\cdots,x_{d_{\text{in}}})^T$.

In many machine learning applications, the goal is to train a neural network using a set of training data $\mathcal{T}_m$. Each datum is a pair of an input and an output, $(\x, \y) \in \mathcal{X}\times \mathcal{Y}$. Here $\mathcal{X} \subset \mathbb{R}^{d_\text{in}}$ is the input space and $\mathcal{Y}  \subset \mathbb{R}^{d_\text{out}}$ is the output space.
Thus, we write $\mathcal{T}_m = \{(\x_i, \y_i)\}_{i=1}^m$.
In order to measure the discrepancy between a prediction and an output,
we introduce a loss metric $\ell(\cdot,\cdot):\mathcal{Y}\times \mathcal{Y} \mapsto \mathbb{R}$ to define a loss function $\mathcal{L}$:
\begin{equation} \label{def:loss}
	\mathcal{L}(\bm{\theta}) = \frac{1}{m}\sum_{i=1}^m \ell(\N^L(\x_i;\bm{\theta}), \y_i).
\end{equation}
For example, the squared loss $\ell(\hat{\y},\y) = \|\hat{\y}-\y\|^2$, logistic $\ell(\hat{y},y) = \log(1+\exp(-y\hat{y}))$, hinge, or cross-entropy are commonly employed.
We then seek to find $\bm{\theta}^*$ which minimizes the loss function $\mathcal{L}$.
In general, a gradient-based optimization method is employed for the training.
In its very basic form, given an initial value of $\bm{\theta}^{(0)}$, the parameters are updated according to
$$
\bm{\theta}^{(k+1)} = \bm{\theta}^{(k)} - \eta_k \frac{\partial \mathcal{L}(\bm{\theta})}{\partial \bm{\theta}}\bigg|_{\bm{\theta} = \bm{\theta}^{(k)}}, 
$$
where $\eta_k$ is the learning rate of the $k$-th iteration.

\subsection{Weights and Biases Initialization and Data Normalization}
Gradient-based optimization is a popular choice for training a neural network.
It commences with the weight and bias initialization.
How to initialize the network plays a crucial role in the success of the training.
Typically, the weights are randomly initialized from probability distributions.  
However, the biases could be set to zeros initially or could be randomly initialized.

In this paper, we consider the following 
weights and biases initialization schemes.
One is the normal initialization. 
That is, all weights and/or biases in the $(t+1)$-th layer are independently initialized from zero-mean normal distributions. 
\begin{equation} \label{initialization-He}
    \begin{split}
        (\text{`Normal' without bias})& \quad \bm{W}^{t+1}_j \sim N\left(0,\sigma_{t+1}^2\bm{I}_{n_{t}}\right), \quad \bm{b}^{t+1}_j = 0,\\
    (\text{`Normal' with bias})& \quad \bm{W}^{t+1}_j \sim N\left(0,\sigma_{t+1}^2\bm{I}_{n_{t}}\right), \quad \bm{b}^{t+1}_j \sim N(0,\sigma_{b,t+1}^2),
    \end{split}
\end{equation}
where $\bm{I}_m$ is the identity matrix of size $m \times m$.
When $\sigma_{t+1}^2 = \frac{2}{n_t}$
and $\bm{b}_j^{t+1} = 0$, 
the initialization is known as the `He initialization' \citep{he2015delving}. 
The `He initialization' is one of the most popular initialization methods for ReLU networks. 
The other initialization is from the uniform distribution on 
the unit hypersphere.
That is, each row of either $\bm{V}^{t+1}$ or $\bm{W}^{t+1}$ is independently initialized from its corresponding the unit hypersphere uniform distribution. 
\begin{equation}\label{initialization-nsphere}
    \begin{split}
        (\text{`Unit hypersphere' without bias})& \quad \bm{W}^{t+1}_j \sim \text{Unif}(\mathbb{S}^{n_{t}-1}), \quad \bm{b}^{t+1}_j = 0,\\
    (\text{`Unit hypersphere' with bias})& \quad \bm{V}^{t+1}_j=[\bm{W}^{t+1}_j, \bm{b}^{t+1}_j] \sim \text{Unif}(\mathbb{S}^{n_{t}}).
    \end{split}
\end{equation}
Throughout this paper, we assume that the training input domain is the closed ball with radius $r>0$, i.e.,
$
B_r(0) = \{\x \in \mathbb{R}^{d_\text{in}} | \|\x\|_2 \le r\}.
$
In many practical applications such as image processing or classification, 
there is a natural bound on the magnitude of each datum.
Also, in practice, the training data is often normalized to have mean zero and variance 1.
Given a training data set $\mathcal{T}_m = \{(\x_i,\y_i)\}_{i=1}^m$, 
the normalization makes $\|\x_i\|_2^2 \le 1$ for all $i=1,\cdots,m$.
Thus, one may assume that the training input data domain is the unit closed ball. 
We note that this assumption is independent of the actual data domain.
This is because one can normalize the given data set since we \textit{always} have a finitely many data.
Many theoretical works \citep{allen2018convergence, du2018gradient-DNN, li2018learning, soltanolkotabi2019theoretical, zou2018stochastic} 
also assume a certain data normalization: Given a training data $\{\x_i\}$,
by letting $\z_i=[\x_i; 1]$, 
$\{\z_i\}$ is normalized to have a unit norm. 
For example, if $\|\x_i\|=r$, its corresponding normalized $\z_i$ is $[\frac{\x_i}{\sqrt{k_ir^2}} ; \frac{\sqrt{k_i-1}}{\sqrt{k_i}}]$
for any $k_i > 1$.
In \cite{allen2018convergence}, $k_i$ was chosen to be $2$ for all $i$.
To this end, $\x_i$ is normalized to $\frac{\x_i}{\sqrt{k_i r^2}}$.
In the later sections, we will see that the choice of $k_i$ will affect the trainability of ReLU networks.

\subsection{Dying ReLU and Born Dead Probability}
\label{subsec:dying ReLU}
Dying ReLU refers to the problem when ReLU neurons become inactive and only output a constant for any input.
We say that a ReLU neuron in the $t$-th hidden layer is dead on $B_r(0)$ if it is a constant function on $B_r(0)$. That is, 
there exists a constant $c \in \mathbb{R}^{+} \cup \{0\}$ such that
$$
\phi(\bm{w}^T\phi(\N^{t-1}(\x)) + b) = c,
\quad \forall \x \in B_r(0).
$$
Also, a ReLU neuron is said to be born dead (BD) if it is dead at the initialization. 
In contrast, a ReLU neuron is said to be active in $B_r(0)$ if it is not a constant function on $B_r(0)$. 
The notion of born death was introduced in \citep{lu2019dying}, where a ReLU network is said to be BD if there exists a layer where all neurons are BD.
We refer to the probability that a ReLU neuron is BD
as the born dead probability (BDP) of a ReLU neuron.

In the 1st hidden layer, once a ReLU neuron is dead, it cannot be revived during the training.
However, a dead neuron in the $t$-th layer where $t>1$ could be revived by other active neurons in the same layer.
In the following, we provide a condition on which a dead neuron cannot be revived.
The lemma is based on Lemma 10 of \cite{lu2019dying}.
\begin{lemma} \label{THM:DEAD-STAY-DEAD} 
    For a shallow ReLU network ($L=2$), none of the dead neurons can be revived through gradient-based training.
    For a deep ReLU network ($L>2$),
    suppose the weight matrices are initialized from probability distributions, which satisfy 
    $\text{Pr}(\bm{W}^t_j\bm{z} = 0) = 0$ for any nonzero vector $\bm{z}$.
    If there exists a hidden layer whose neurons are all dead,
    with probability 1,
    none of the dead neurons can be revived through gradient-based training. 
\end{lemma}
\begin{proof}
	The proof can be found in \ref{app:thm:dead-stay-dead}.
\end{proof}

\section{Trainability of ReLU Networks}
\label{sec:trainability}

\subsection{Shallow ReLU Networks}
For pedagogical reasons, we first confine ourselves to shallow (1-hidden layer) ReLU networks. 
For shallow ReLU networks, we define the trainability as follows:
\begin{definition} \label{def:trainability}
	For a learning task that requires at least $m$ active neurons,
	a shallow ReLU network of width $n$ is said to be trainable if the number of active neurons is greater than or equal to $m$.
	We refer to the probability of a network being trainable at the initialization as
	trainability.
\end{definition}
From Lemma~\ref{THM:DEAD-STAY-DEAD}, dead neurons will never be revived during the training.
Thus, given a learning task which requires at least $m$ active neurons,
in order for successful training,
an initialized network should have at least $m$ active neurons in the first place.
If the number of active neurons is less than $m$, there is no hope to train the network successfully. 
Therefore, \textit{a network being trainable is a necessary condition for successful training}. 
We note that this condition is independent of the choice of loss metric $\ell(\cdot,\cdot)$ in \eqref{def:loss},
of the number of training data, and of the choice of gradient-based optimization methods.

We now present the trainability results for shallow ReLU networks.
\begin{theorem} \label{THM:MAIN}
    Given a learning task, which requires a shallow ReLU network having at least $m$ active neurons,
    suppose the training input domain is $B_r(0)$
    and a shallow network of width $n \ge m$ is employed.
    \begin{itemize}[leftmargin=*]
        \item If either the `normal' \eqref{initialization-He} or the `unit hypersphere' \eqref{initialization-nsphere} 
        initialization without bias is used in the 1st hidden layer,
	with probability 1, the network is trainable.
	    \item If either the `normal' \eqref{initialization-He} or the `unit hypersphere' \eqref{initialization-nsphere} 
        initialization with bias is used in the 1st hidden layer,
	    with probability,
	    $$
	    \text{Pr}(\mathfrak{m}_1 \ge m) = \sum_{j=m}^{n} \binom{n}{j}(1-\hat{p}_{d_\text{in}}(r))^{j}(\hat{p}_{d_\text{in}}(r))^{n_1-j},
	    $$
	    where
	    $\hat{p}_d(r)
	    = \frac{1}{\sqrt{\pi}}\frac{\Gamma((d+1)/2)}{\Gamma(d/2)}\int_0^{\alpha_r} (\sin u)^{d-1} du$ and
	    where $\alpha_r = \tan^{-1}(1/r)$,
        the network is trainable.
        Furthermore, on average, at least 
        $$
        n\left(1-\sqrt{\frac{d_\text{in}}{2\pi}}\alpha_r (\sin \alpha_r)^{d_\text{in}-1}\right)
        $$
        neurons will be active at the initialization.
    \end{itemize}
\end{theorem}
\begin{proof}
	The proof can be found in \ref{app:THM:MAIN}.
\end{proof}
Theorem~\ref{THM:MAIN} implies that if the biases are randomly initialized, over-specification is necessary for successful training.
It also shows a degree of over-specification whenever one has a specific width in mind for a learning task.
If it is known (either theoretically or empirically) that a shallow network of width $m$ can achieve a good performance, 
one should use a network of width 
$n = \frac{m}{1 - \hat{p}_{d_\text{in}}(r)}$
to guarantee that the initialized network has $m$ active neurons (on average) at the initialization. 
For example, when $d_\text{in}=1$, $r=1/\sqrt{3}$, $m=200$, it is suggested to work on a network of width $n= 300$ in the first place. 
The example (Figure~\ref{fig:motivation}) given in section~\ref{sec:introduction} can be understood in this manner.
By Theorem~\ref{THM:MAIN}, with probability at least 0.43, 
the network of width 2 fails to be trained successfully
for any learning task that requires at least 2 active neurons.
The trainability depends only on $\hat{p}_d(r)$, which evidently shows its dependency on the maximum magnitude $r$ of training data.
The smaller the $r$ is, the larger the $\hat{p}(r)$ becomes.
This indicates that how the data are normalized also affects the trainability. 

On the other hands, if the biases are initialized to zero,
over-parameterization or over-specification is not needed from this perspective. 
However, the zero-bias initialization often finds a spurious local minimum
or gets stuck on a flat plateau.
In section~\ref{sec:data-dependent}, we further investigate the bias initialization.

Next, we provide two concrete learning tasks that require a certain number of active neurons. 
For this purpose, we introduce the minimal function class.
\begin{definition} \label{def:minimal-func-class}
	Let $\mathcal{F}_n(r)$ be a class of shallow ReLU neural networks of width $n$ defined on $B_r(0)$;
	\begin{equation*} 
	\mathcal{F}_n(r) = \left\{\sum_{i=1}^n c_i \phi(\bm{w}_i^T\x + b_i) + c_0 \hspace{0.1cm} \bigg| \forall i, \phi(\bm{w}_i^T\x+b_i) \text{ is active in } B_r(0) \right\},
	\end{equation*} 
	where $c_i, b_i \in \mathbb{R}$, $c_i \ne 0$, and  $\bm{w}_i \in \mathbb{R}^{d_\text{in}}$ for $i=1,\cdots,n$.
	Given a continuous function $f$ and $\epsilon > 0$, a function class $\mathcal{F}_{m_\epsilon}$ is said to be the $\epsilon$-minimal function class for $f$ 
	if $m_\epsilon$ is the smallest number such that $\exists g \in \mathcal{F}_{m_\epsilon}(r)$ and $|g-f| < \epsilon$ in $B_r(0)$.
	If $\epsilon = 0$, we say $\mathcal{F}_{m_0}(r)$ is the minimal function class for $f$. 
\end{definition}
We note that $\mathcal{F}_j \cap  \mathcal{F}_{s} = \emptyset$ for $j \ne s$,
and a function $f \in \mathcal{F}_j(r)$ could allow different representations in other function classes $\mathcal{F}_s(r)$ for $s > j$ in $B_r(0)$. 
For example, $f(x)=x$ on $B_r(0)=[-r,r]$ can be expressed as either 
$g_1(x) = \phi(x+r) -r \in \mathcal{F}_1(r)$,
or
$g_2(x) = \phi(x) - \phi(-x) \in \mathcal{F}_2(r)$.
However, it cannot be represented by $\mathcal{F}_0(r)$.
Thus, $\mathcal{F}_1(r)$ is the minimal function class for $f(x)=x$.
We remark that $g_1$ and $g_2$ are not the same function in $\mathbb{R}$, however, they are the same on $B_r(0)$.
Also, note that the existence of $m_\epsilon$ in Definition~\ref{def:minimal-func-class} is guaranteed by 
universal function approximation theorems for shallow neural networks \citep{hornik1991approximation, cybenko1989approximation}.
Hence, approximating a function whose minimal function class is $\mathcal{F}_m(r)$
is a learning task that requires at least $m$ active neurons.
Also, we say any ReLU network of width greater than $m_\epsilon$ to be over-specified for approximating $f$ within $\epsilon$.

A network is said to be over-parameterized if the number of parameters is larger than the number of training data.
In this paper, we consider the over-parameterization, where the size of width is greater than or equal to the number of training data.
Then, over-parameterization can be understood under the frame of over-specification
by the following lemma.
\begin{lemma} \label{THM:INTERPOLATION}
	For any non-degenerate $(m+1)$ training data, there exists a shallow ReLU network of width $m$ which interpolates
	all the training data.
	Furthermore, there exists non-degenerate $(m+1)$ training data such that any shallow ReLU network of width less than $m$ cannot interpolate all the training data.
	In this sense, $m$ is the minimal width.
\end{lemma}
\begin{proof}
	The proof can be found in \ref{app:THM:INTERPOLATION}.
\end{proof}
Lemma~\ref{THM:INTERPOLATION} shows that 
any network of width greater than $m$
is over-specified for interpolating $(m+1)$ training data.
Thus, we could regard over-parameterization as a kind of over-specification.
Hence, interpolating any non-degenerate $(m+1)$ training data is also a learning task that requires at least $m$ active neurons.

With the trainability obtained in Theorem~\ref{THM:MAIN}, 
we show that over-parameterization is both a necessary and a sufficient condition for minimizing the loss.
\begin{theorem} \label{THM:OVERPARA-IFF}
     For shallow ReLU networks,
     suppose either the `normal' \eqref{initialization-He} or the `unit hypersphere' \eqref{initialization-nsphere} initialization with bias is employed in the first hidden layer.
     Also, the training input domain 
     is $B_r(0)$.
     For any non-degenerate $(m+1)$ training data, which requires a network to have at least $m$ active neurons
     for the interpolation, 
     suppose $m$ and the input dimension $d_\text{in}$ satisfy
     \begin{equation} \label{eqn:overpara-iff}
         1 - (1-\delta)^{1/m} < \frac{\exp(-C_rd_\text{in})}{\pi d_\text{in}}, \qquad
     C_r = -\log(\sin(\tan^{-1}(1/r))),
     \end{equation}
     where $0 < \delta < 1$.
     Then,
     over-parameterization is both 
     a necessary and a sufficient condition
     for interpolating all the training data with probability at least $1-\delta$ over the random initialization by the (stochastic) gradient-descent method.
\end{theorem}
\begin{proof}
    The proof can be found in \ref{app:THM:OVERPARA-IFF}.
\end{proof}

We remark that Theorem~\ref{THM:OVERPARA-IFF} assumes 
that the biases are randomly initialized.
To the best of our knowledge, all existing theoretical results
also assume the random bias initialization,
e.g. \cite{du2018gradient-shallow, oymak2019towards,li2018learning}.

\subsection{Trainability of Deep ReLU Networks}
We now extend the notion of trainability to 
deep ReLU networks.
Unlike dead ReLU neurons in the 1st hidden layer, 
a dead neuron in the $t$-th hidden layer ($t > 1$)
could be revived during the training if two conditions are satisfied. 
One is that for all layers, there exists at least one active neuron.
This condition is directly obtained from Lemma~\ref{THM:DEAD-STAY-DEAD}.
The other is that the dead neuron should be in the condition of \textit{tentative death}, that will be introduced shortly.
We remark that these two conditions are necessary conditions for the revival of a dead neuron.
We now provide a precise meaning of the tentative death as follows. 

Let us consider a neuron in the $t$-th hidden layer;
$$
\phi(\bm{w}^T\x^{t-1} + b), \qquad 
\x^{t-1} = \phi(\N^{t-1}(\x)).
$$
Suppose the neuron is dead.
For any changes in $\x^{t-1}$, but not in  $\bm{w}$ and $b$, 
if the neuron is still dead,
we say a neuron is \textit{permanently dead}.
For example, if $\bm{w}_j, b \le 0$ and $t>1$, since $\x^{t-1} \ge 0$, regardless of how $\x^{t-1}$ changes,
the neuron will never be active again. 
Hence, in this case, there is no hope that the neuron can be revived during the gradient training. 
Otherwise, we say a neuron is \textit{tentatively dead}.
Therefore, any neuron is always in one of three states: active, tentatively dead, and permanently dead.

We now define the trainability for deep ReLU networks.
\begin{definition} \label{def:trainability-DL}
	For a learning task that requires a $L$-layer ReLU network having at least $m_t$ active neurons in the $t$-th layer,
	a $L$-layer ReLU network with $\bm{n}=(n_0,n_1,\cdots,n_L)$ architecture
	is said to be trainable 
	if the number of permanently dead neurons in the $t$-th layer is less than or equal to $n_t - m_t$ for all $1 \le t < L$.
	We refer to the probability of a network being trainable at the initialization as trainability.
\end{definition}
For $L=2$, since there is no tentatively dead neuron,
Definition~\ref{def:trainability} becomes a special case of Definition~\ref{def:trainability-DL}.


We now present the trainability results for ReLU networks of depth $L=3$ at $d_\text{in}=1$.
Since each layer can be initialized in different ways, 
we consider some combinations of them.
\begin{theorem} \label{THM:DEEP}
	Suppose the training input domain is $B_r(0)$ and $d_\text{in} = 1$.
	For a learning task that requires a 3-layer ReLU network having at least $m_t$ active neurons in the $t$-th layer,
	a 3-layer ReLU network with $\bm{n}=(1,n_1,n_2,n_3)$ architecture is initialized as follows.
	(Here $n_1 \ge m_1$, $n_2 \ge m_2$ and $n_3 = m_3$):
	\begin{itemize}[leftmargin=*]
		\item Suppose the `unit hypersphere' \eqref{initialization-nsphere} initialization without bias is used in the 1st hidden layer.
		\begin{enumerate}
			\item If the `normal' \eqref{initialization-He} initialization without bias is used in the 2nd hidden layer, 
			with probability at least, 
			$$
			\sum_{j=m_2}^{n_2} \binom{n_2}{j}\left[\left(1-\frac{1}{2^{n_1-1}}\right)\frac{3^j}{4^{n_2}} + \frac{1}{2^{n_1+n_2-1}}\right]
			+ \sum_{j=1}^{m_2-1}\sum_{l=0}^{n_2-m_2}
			\binom{n_2}{n_2-j-l,j,l} Q(j,l)
			$$
			where
			$$
			Q(j,l)=
			\left[\frac{(1-2^{-n_1})^{n_2-j-l}}{2^{(l+1)n_1+n_2-1}}
			+ \left(1 - \frac{1}{2^{n_1-1}}\right)\frac{3^j(3/4-2^{-n_1-1})^{n_2-j-l}}{2^{(n_1+1)l+2j}}
			\right],
			$$
			the network is trainable.
			\item If the `normal' \eqref{initialization-He} initialization with bias is used in the 2nd hidden layers, 
			with probability at least, 
			$$
			(1-\hat{p}_d(r))^{n_1}\left[\sum_{j=m_2}^{n_2} \binom{n_2}{j}\hat{Q}(j) + \sum_{j=1}^{m_2-1}\sum_{l=0}^{n_2-m_2}
				\binom{n_2}{n_2-j-l,j,l} Q(j,l)\right],
			$$
			where $\hat{p}_d(r)$ is defined in Theorem~\ref{THM:MAIN}, $s \sim B(n_1,1/2)$, $\alpha_s = \tan^{-1}(\frac{s}{n_1-s})$, $g(x) = \sin(\tan^{-1}(x))$, and
			\begin{align*}
			\hat{Q}(j) &=\mathbb{E}_{s}\left[(1-\mathfrak{p}_2(s))^j\mathfrak{p}_2(s)^{n_2-j}\right], \\
			\mathfrak{p}_2(s) &= \frac{1}{2} + \left[\int_{\frac{\pi}{2}}^{\pi+\alpha_{s}} \frac{g(r\sqrt{s}\cos(\theta))}{4\pi}d\theta + \int_{\pi+\alpha_{s}}^{2\pi} \frac{g(r\sqrt{n_1-s}\sin(\theta))}{4\pi}d\theta \right],
			\\
			Q(j,l) &=
				\mathbb{E}_{s}\left[(1-\mathfrak{p}_2(s))^{j}
							(\mathfrak{p}_2(s) - 2^{-n_1-1})^{n_2-j-l}
							(2^{-n_1-1})^{l}\right],
			\end{align*}
			the network is trainable.
		\end{enumerate}
		\item Suppose the `unit hypersphere' \eqref{initialization-nsphere} initialization with bias is used in the 1st hidden layer. 
		\begin{enumerate}
			\item If the `normal' \eqref{initialization-He} initialization without bias is used in the 2nd hidden layer and $n_1=m_1=1$,
			with probability at least, 
			$$
			2^{-n_2}\sum_{j=m_2}^{n_2} \binom{n_2}{j} + \sum_{j=1}^{m_2-1}\sum_{l=0}^{n_2-m_2} 
			\binom{n_2}{n_2-j-l,j,l} 2^{-2n_2+j},
			$$
			where
			the network is trainable.
			\item If the `normal' \eqref{initialization-He} initialization with bias is used in the 2nd hidden layers and $n_1=m_1=1$, 
			with probability at least, 
			\begin{equation*}
			(1-\hat{p}_d(r))^{n_1}\left[\sum_{j=m_2}^{n_2} \binom{n_2}{j}\mathbb{E}_{\omega}\left[(1-\mathfrak{p}_2(\omega))^j\mathfrak{p}_2(\omega)^{n_2-j}\right] + Q\right],
			\end{equation*}
			where $\hat{p}_d(r)$ is defined in Theorem~\ref{THM:MAIN},
			$\alpha_r = \tan^{-1}(r)$, $\omega \sim \text{Unif}\left(0,\frac{\pi}{2}+\alpha_r\right)$,
			$g(x) = \tan^{-1}(\frac{1}{\sqrt{r^2+1}\cos(x)})$,
			and
			\begin{align*}
			\mathfrak{p}_2(\omega) &=
			\begin{cases}
			\frac{1}{4} + \frac{g(\omega - \alpha_r)}{2\pi}, 
			&\text{if } 
			\omega \in \left[\frac{\pi}{2}-\alpha_r, \frac{\pi}{2}+\alpha_r\right), \\
			\frac{1}{4} +\frac{g(\omega - \alpha_r) + \tan^{-1}\left(\sqrt{r^2+1}\cos(\omega + \alpha_r)\right)}{2\pi},
			& \text{if }
			\omega \in \left[0, \frac{\pi}{2}-\alpha_r\right),
			\end{cases}, \\
			Q&=
			\sum_{j=1}^{m_2-1}\sum_{l=0}^{n_2-m_2}
			\binom{n_2}{n_2-j-l,j,l} \mathbb{E}_{\omega}\left[(1-\mathfrak{p}_2(\omega))^{j}
				(\mathfrak{p}_2(\omega) - 2^{-2})^{n_2-j-l}
				(2^{-2})^{l}
				\right],
			\end{align*}
			the network is trainable.
		\end{enumerate}
	\end{itemize}
\end{theorem}
\begin{proof}
	The proof can be found in ~\ref{sec:gen-formula-trainability}.
\end{proof}

Theorem~\ref{THM:DEEP} suggests us to use a ReLU network with sufficiently large width at each layer to secure a high trainability.
Also, it is clear that different initialization schemes result in different trainabilities.
Our proof is built on the study of the probability distribution of the number of active neurons (see Lemma~\ref{THM:PI_2}).
In Figure~\ref{fig:pi_j-1642} of ~\ref{sec:analysis}, 
we illustrate the active neuron distributions
by three different initialization schemes.

At last, we present an upper bound of the trainability when the biases are initialized to zeros. 
\begin{corollary} \label{COR:TRAINABILITY-UPPER-1D}
    For a learning task that requires a $L$-layer ReLU network having at least $m_t$ active neurons in the $t$-th layer,
    suppose that all weights are independently initialized from the `normal' \eqref{initialization-He} initialization without bias, and $d_\text{in}=1$. 
    Then, 
    the trainability of a $L$-hidden layer ReLU network having $n \ge m_t$ neurons at each layer
    is bounded above by 
    \begin{equation*}
         \mathfrak{a}_1^{L-1} -
        \frac{(1-2^{-n+1})(1-2^{-n})}{1+(n-1)2^{-n}}(-\mathfrak{a}_1^{L-1} + \mathfrak{a}_2^{L-1}),
    \end{equation*}
    where
    $\mathfrak{a}_1 = 1-2^{-n}$
    and $\mathfrak{a}_2 = 1-2^{-n+1}-(n-1)2^{-2n}$.
\end{corollary}
\begin{proof}
    The proof can be found in \ref{app:COR:TRAINABILITY-UPPER-1D}.
\end{proof}

Further characterization will be deferred to a future study, however, 
a general formulation is established and can be found in Lemma~\ref{lem:gen-formula-trainability} in \ref{sec:gen-formula-trainability} 
for the readability.

In principle, a single active neuron in the highest layer could potentially revive  tentatively dead neurons through back-propagation (gradient).
However, in practice, it would be better an initialized network to have at least $m_t$ active neurons in the $t$-th hidden layer
for both faster training and robustness.
Let $A$ be the event that a ReLU network has at least $m_t$ active neurons in the $t$-th hidden layer for $t=1,\cdots,L$.
The probability of $A$ is then a naive lower bound of trainability.
Hence, having a high probability of $A$ enforces a high trainability.

\textbf{Remark:} A trainable network itself does not guarantee successful training.
However, if a network is not trainable, 
there is no hope for the network to be trained successfully.
Thus, \textit{a network being trainable is
a necessary condition for successful training}.
And the trainability
serves as an upper bound of the training success rate.
The demonstration of trainability is given in section~\ref{sec:example}.

\section{Data-dependent Bias Initialization: Shallow ReLU Networks}
\label{sec:data-dependent}
\begin{figure}[!htbp] 
	\centerline{
		\includegraphics[width=7cm]{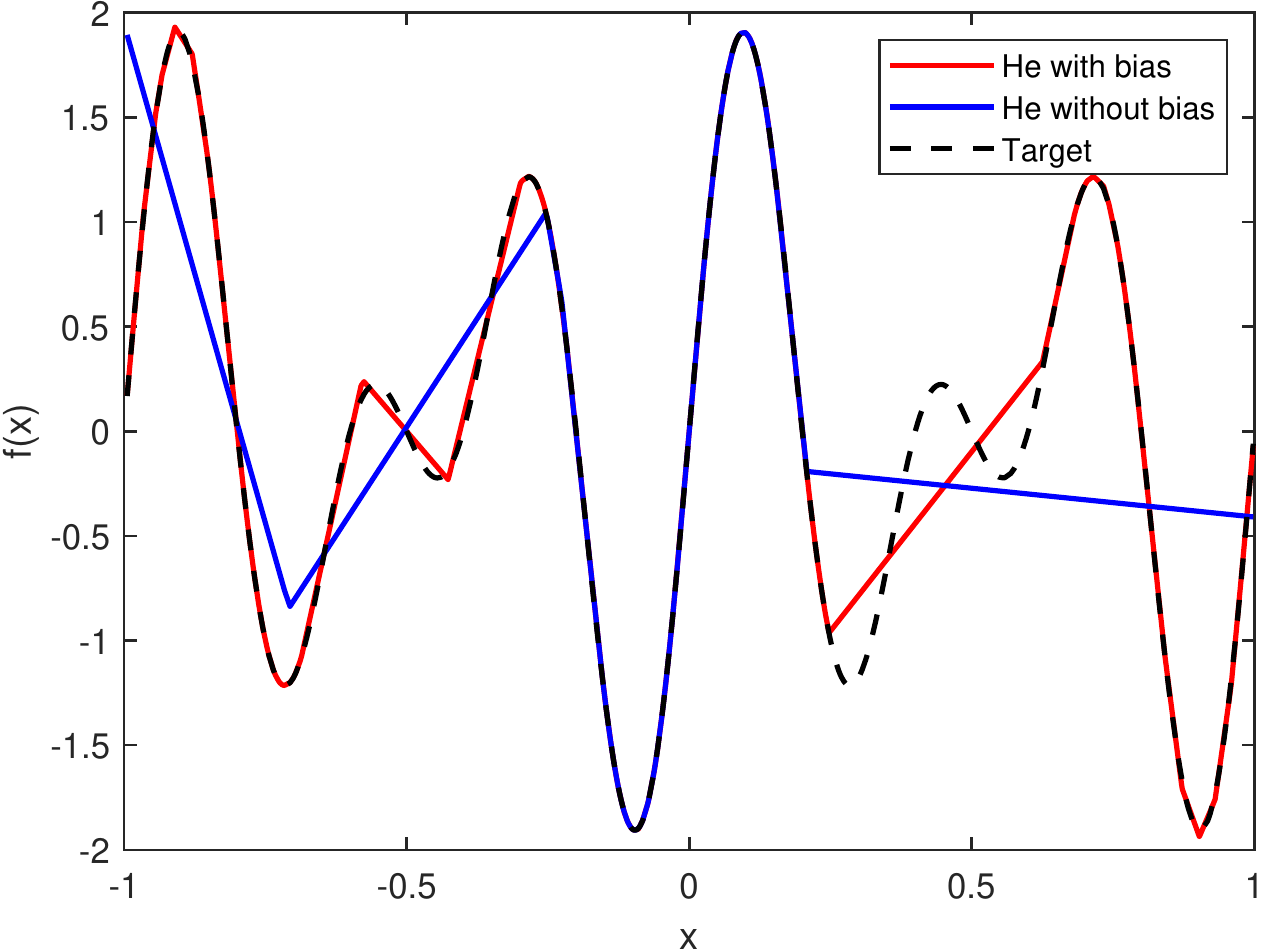}
	}
	\caption{The trained networks for approximating $f(x)=\sin(4\pi x) + \sin(6\pi x)$ 
	by the `He initialization' without bias and with bias.
	A shallow ReLU network of width 500 is employed.
	The target function $f(x)$ is also plotted.}
	\label{fig:demo-sin2pi}
\end{figure}

In this section, we investigate the bias initialization in the gradient-based training. 
In terms of trainability for shallow ReLU networks,
Theorem~\ref{THM:MAIN} indicates that 
the zero-bias initialization would be preferred over the random bias initialization.
In practice, however, 
the zero-bias initialization 
often finds a spurious local minimum
or gets stuck on a flat plateau.
To illustrate this difficulty, we consider a problem of approximating a sum of two sine functions $f(x) = \sin(4\pi x) + \sin(6\pi x)$ on $[-1,1]$. 
For this task, we use a shallow ReLU network of width 500 with the `He initialization' without bias.
In order to reduce extra randomness in the experiment, 
100 equidistant points on $[-1,1]$ are used as the training data set.
One of the most popular gradient-based optimization methods, \texttt{Adam} \citep{kingma2014adam}, is employed 
with its default parameters. We use the full-batch size and set the maximum number of epochs to 15,000.
The trained network is plotted in 
Figure~\ref{fig:demo-sin2pi}.
It is clear that 
the trained network is stuck on a local minimum. 
A similar behavior is repeatedly observed 
in all of our multiple independent simulations. 

This phenomenon could be understood as follows. 
Since the biases are zero, all initialized neurons are clustered at the origin.
Consequently, it would take long time for gradient-update to distribute neurons over the training domain to achieve a small training loss. In the worst case, along the way of distributing neurons, it will find a spurious local minimum.
We refer to this problem as the \textit{clustered neuron problem}.
Indeed, this is observed in Figure~\ref{fig:demo-sin2pi}.
The trained network well approximates the target function on a small domain containing the origin, however, 
it loses its accuracy on the domain far from the origin. 

On the other hand, if we randomly initialize the bias, as shown in Theorem~\ref{THM:MAIN}, over-specification is inevitable to guarantee a certain number of active neurons.
In this setting, at the initialization, only 375 neurons will be active among 500 neurons on average.
In Figure~\ref{fig:demo-sin2pi}, we also show the trained result by the `He initialization' with bias.
Since neurons are now randomly distributed over the entire domain,
the trained network approximates quite well the target function.
However, the randomness may locate some neurons in places that may lead to a spurious local minimum or a slow training.
In the worst case, some neurons would never be activated. 
In this example, the trained network by the random bias initialization 
loses its accuracy at some parts of the domain, e.g. in the intervals containing $\pm0.5$.

In order to overcome such difficulties and accelerate the gradient-based training, 
we propose a new data-dependent initialization scheme. 
The scheme is for the over-parameterized setting, where the size of width is greater than or equal to the number of training data. 
By adapting the trainability perspective, 
the method is designed to alleviate both the clustered neuron problem and the dying ReLU neuron problem at the same time.
This is done by efficiently locating each neuron based on the training data.

\textbf{Remark:} We aim to study the effect of bias initialization on the gradient-based training.
Interpolating all the training data results in the zero training loss.
However, we do not simply attempt to interpolate the training data, which can be done by explicit construction shown in Lemma~\ref{THM:INTERPOLATION}.
We remark that the idea of data-dependent initialization is not new; see  \citep{ioffe2015batch, krahenbuhl2015data, salimans2016weight}.
However, our method is specialized to the over-parameterized setting.


\subsection{Data-dependent Bias Initialization}
Let $m$ be the number of training data
and $n$ be the width of a shallow ReLU network. 
Suppose the network is over-parameterized so that $n = hm$ for some positive number $h \ge 1$. 
We then propose to initialize the biases as follows;
\begin{align*}
\bm{b}_i = -\bm{w}_i^T\x_{j_i} + |\epsilon_{i}|, \qquad \epsilon_i \sim N(0,\sigma_{e}^2),
\end{align*}
where $\epsilon_i$'s are iid
and $j_i -1 = (i-1) \mod m$.
We note that this mimics the explicit construction for the data interpolation
in Lemma~\ref{THM:INTERPOLATION}.
By doing so, the $i$-th neuron is initialized to be located near $\x_{j_i}$ as
$$
\phi(\bm{w}^T_i(\x - \x_{j_i}) + |\epsilon_i|).
$$
The precise value of $\sigma_{e}^2$ is determined as follows.
Let $q(\x)$ be the expectation of the normalized squared norm of the network, i.e.,
$
q(\x):=\mathbb{E}[\|\N(\x)\|_2^2]/d_\text{out},
$
where the expectation is taken over weights and biases
and $\N(\x)$ is a shallow ReLU network having $\bm{n}=(d_\text{in},n,d_\text{out})$ architecture.
Given a set of training input data $\mathcal{X}_m=\{\x_i\}_{i=1}^m$, we define the average of $q(\x)$ on $\mathcal{X}_m$ as 
$$
\mathbb{E}_{\mathcal{X}_m}[q(\x)]:= \frac{1}{m}\sum_{i=1}^m q(\x_i).
$$
We then choose our parameters to match $\mathbb{E}_{\mathcal{X}_m}[q(\x)]$ by our data-dependent initialization to 
the one by the standard initialization method. 
For example, when the `normal' \eqref{initialization-He} initialization without bias is used, we have
$$
\mathbb{E}_{\mathcal{X}_m}[q(\x)]:= \frac{n\sigma_{\text{out}}^2 \sigma_{\text{in}}^2}{2m}\|\bm{X}\|_F^2, \qquad
\bm{X} = [\x_1,\cdots,\x_m],
$$
where $\bm{W}_j^1 \sim N(0,\sigma_{\text{in}}^2\bm{I}_{d_\text{in}})$ for $1\le j \le n$,
    $\bm{W}_i^2 \sim N(0,\sigma_{\text{out}}^2\bm{I}_{n})$ for $1 \le i \le d_\text{out}$,
and $\|\cdot\|_F$ is the Frobenius norm.
When the `He initialization' without bias is used, i.e.,  $\sigma_\text{in}^2 = 2/d_\text{in}$ and $\sigma_{\text{out}}^2 = 2/n$,
we have $\mathbb{E}_{\mathcal{X}_m}[q(\x)] = \frac{2}{d_\text{in}m}\|\bm{X}\|_F^2$.

\begin{theorem} \label{THM:DATA-DEPENDENT}
	For a shallow network of width $n$, 
	suppose $n = hm$ for some positive number $h \ge 1$ where $m$ is the number of training data. 
	Let $\mathcal{X}_m =\{\x_i\}_{i=1}^{m}$ be the set of training input data.
	Suppose $\bm{W}_j^1 \sim N(0,\sigma_{\text{in}}^2\bm{I}_{d_\text{in}})$ for $1\le j \le n$,
    $\bm{W}_i^2 \sim N(0,\sigma_{\text{out}}^2\bm{I}_{n})$ for $1 \le i \le d_\text{out}$,
    $\bm{b}^2 = \bm{0}$, 
    and $\bm{b}^1$ is initialized by the proposed method.
	Then, 
	\begin{equation*} \label{def-sigma-e}
	    \mathbb{E}_{\mathcal{X}_m}[q(\x)] = \frac{h\sigma_\text{out}^2\sigma_\text{in}^2}{m\pi}\sum_{k,i=1}^m\left[(s^2+\Delta_{k,i}^2)h(s/\Delta_{k,i})+s\Delta_{k,i} \right],
	\end{equation*}
	where 
	$h(x) = \tan^{-1}(x) + \pi/2$,
	$\Delta_{k,i} = \|\x_k - \x_i\|_2$ and 
	$\sigma_e = \sigma_\text{in} s$.
\end{theorem}
\begin{proof}
    The proof can be found in \ref{app:THM:DATA-DEPENDENT}.
\end{proof}

For example, if we set $\sigma_\text{in}, \sigma_\text{out}$ and $\sigma_e$ to be
\begin{equation} \label{data-dependent-parameter}
    \sigma_\text{in}^2 = \frac{2}{d_\text{in}}, \quad
\sigma_e^2 = 0, \quad
\sigma_\text{out}^2 = \frac{1}{h}\cdot
\frac{\sum_{j} \|\x_j\|^2}{\sum_{k < i} \|\x_k - \x_i \|^2},
\end{equation}
$\mathbb{E}_{\mathcal{X}_m}[q(\x)]$ by the data-dependent initialization is equal to the one by the `He initialization' without bias.

The proposed initialization makes sure that all neurons are equally distributed over the training data points. 
Also, it would make sure that at least one neuron will be activated at a training datum.
By doing so, it would effectively avoid both the clustered neuron problem
and the dying ReLU neuron problem.
Furthermore, it locates all neurons in favor of the training data points with a hope that 
such neuron configuration accelerates the training.

\begin{figure}[!htbp] 
	\centerline{
		\includegraphics[width=6.5cm]{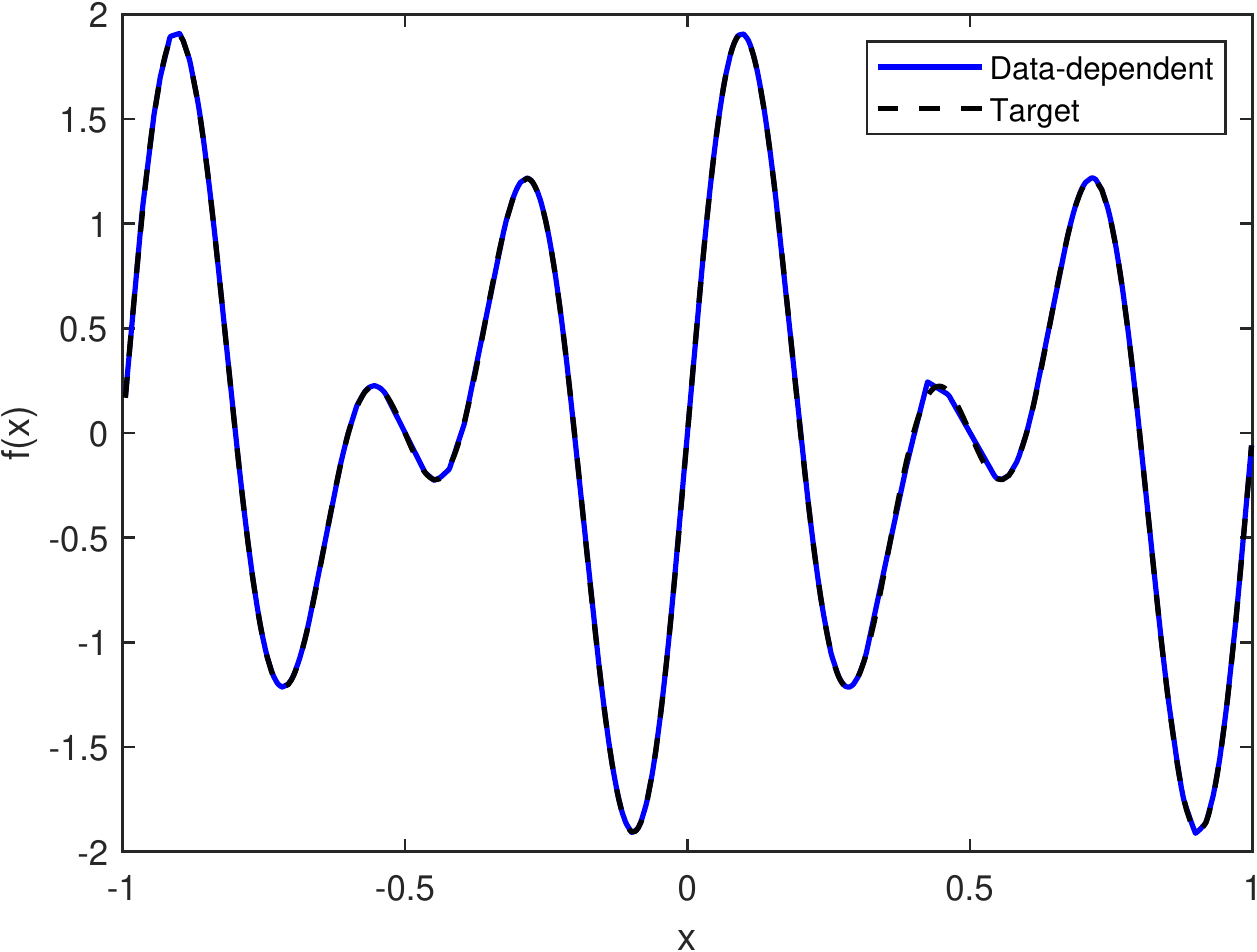}
		\includegraphics[width=6.7cm]{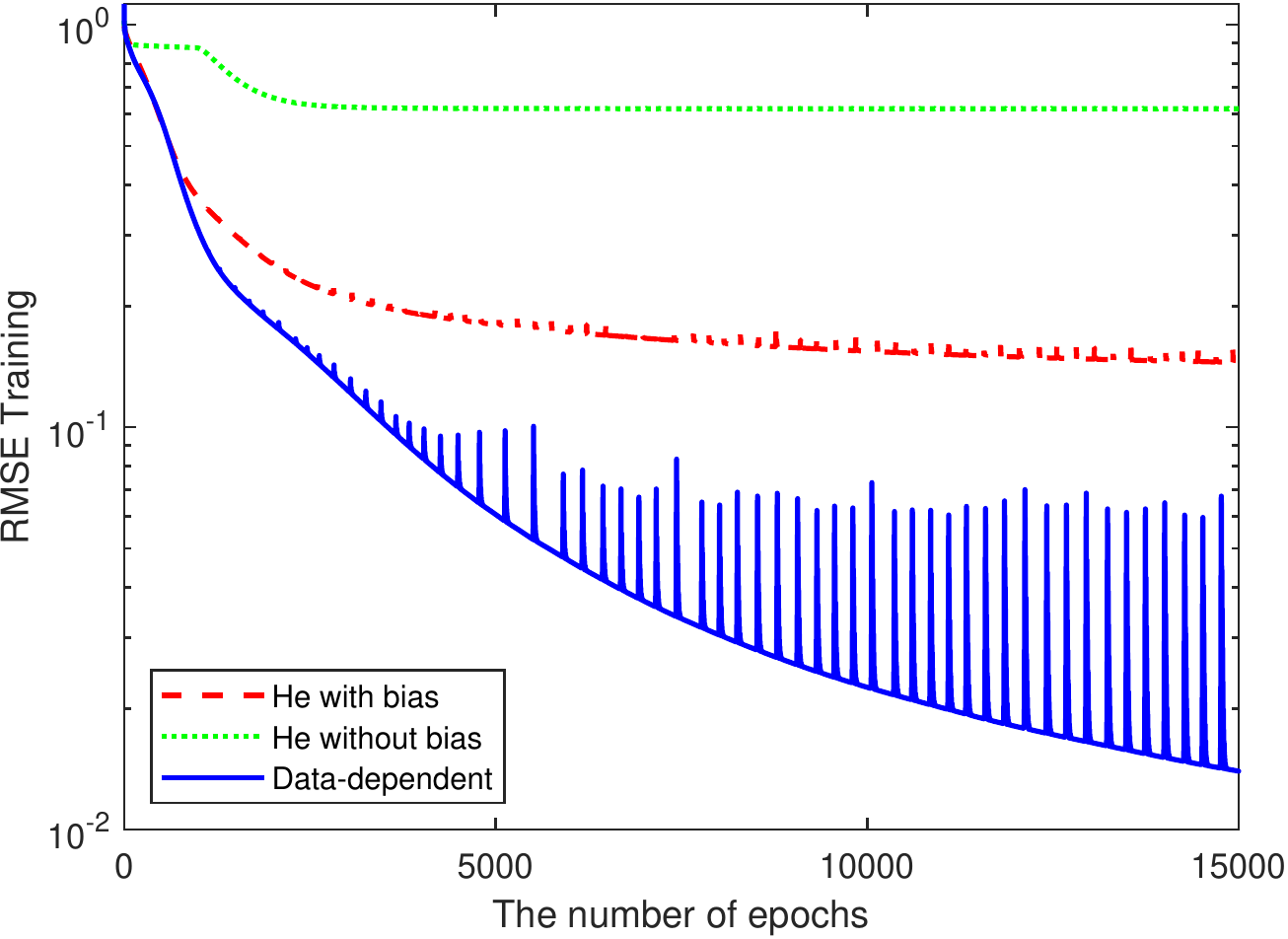}
	}
	\caption{(Left) The trained network for approximating $f(x)=\sin(4\pi x) + \sin(6\pi x)$ by the proposed data-dependent initialization. A shallow ReLU network of width 500 is employed.
	(Right) The root mean square error of the training loss with respect to the number of epochs of \texttt{Adam} \citep{kingma2014adam}.}
	\label{fig:demo-sin4pi-datadep}
\end{figure}

In Figure~\ref{fig:demo-sin4pi-datadep}, we demonstrate the performance of the proposed method in approximating the sum of two sine functions. 
On the left, the trained neural network is plotted,
and on the right the root mean square errors (RMSE) of the training loss are plotted with respect to the number of epochs by three different initialization methods. 
We remark that since the training set is deterministic and the full-batch is used, the only randomness in the training process is from the weights and biases initialization.
It can be seen that the proposed method not only results in 
the fastest convergence but also achieves the smallest approximation error among others.
The number of dead neurons in the trained network 
is 127 (He with bias), 3 (He without bias),
and 17 (Data-dependent). 

\section{Numerical Examples} \label{sec:example}
We present numerical examples to demonstrate our theoretical findings and the effectiveness of the proposed data-dependent initialization method.

\subsection{Trainability of Shallow ReLU Networks}
We present two examples to demonstrate the trainability of a shallow ReLU neural network and justify our theoretical results.
Here all the weights and biases are initialized according to the `He initialization' \eqref{initialization-He} with bias. 
We consider two uni-variate test target functions:
\begin{equation*} 
\begin{split}
f_1(x) &= |x| = \max\{x, 0\} + \max\{-x, 0\}, \\
f_2(x) &=|x| - \frac{\sqrt{3}}{\sqrt{3}-1}\max\{x-1, 0\} -\frac{\sqrt{3}}{\sqrt{3}-1}\max\{-x-1, 0\}.
\end{split}
\end{equation*}
We note that $\mathcal{F}_2$ is the minimal function class (see Definition~\ref{def:minimal-func-class}) for $f_1(x)$ 
and $\mathcal{F}_4$ is the minimal function class for $f_2(x)$. 
That is,
theoretically, $f_1$ and $f_2$ should be exactly recovered by a shallow ReLU network of width $2$ and $4$, respectively.
For the training, we use a training set of 600 data points uniformly generated from $[-\sqrt{3},\sqrt{3}]$ and a test set of 1,000 data points uniformly generated from $[-\sqrt{3},\sqrt{3}]$. 
We employ the standard stochastic gradient descent with mini-batch of size 128 and a constant learning rate of $10^{-3}$.
We set the maximum number of epochs to $10^{6}$ and use the standard square loss.

\begin{figure}[!htbp]
	\centerline{
		\includegraphics[width=5.25cm]{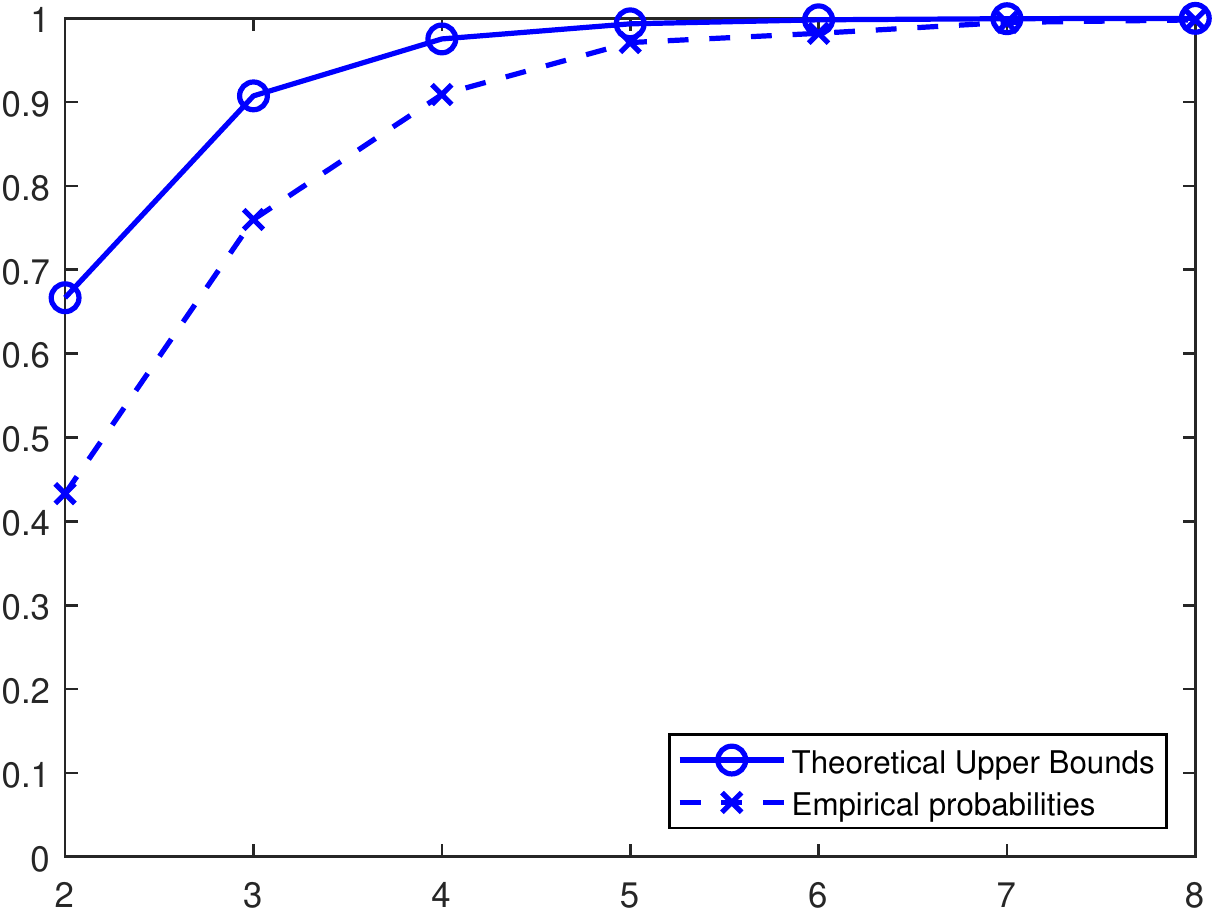}
		\includegraphics[width=4.95cm]{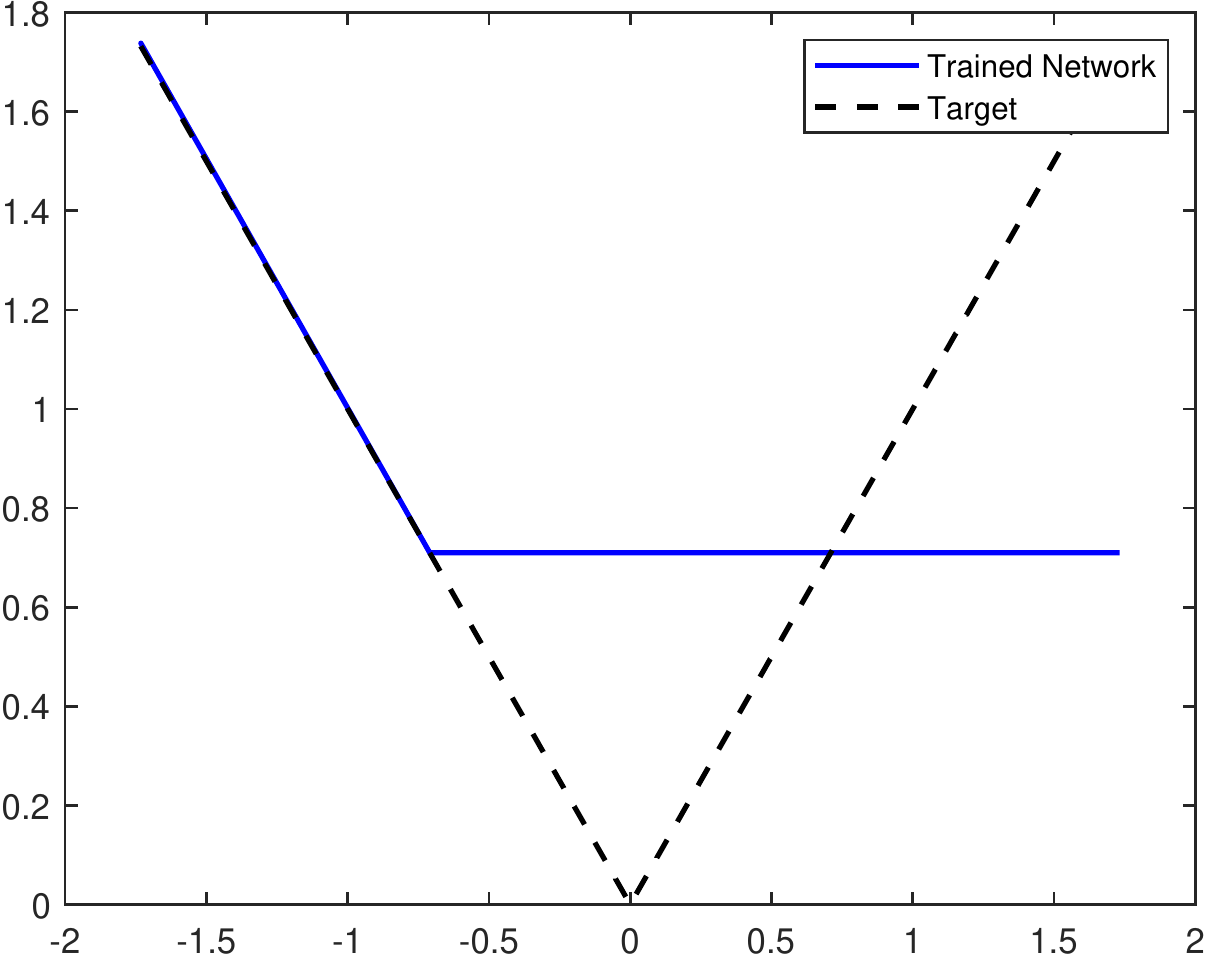}
		\includegraphics[width=4.95cm]{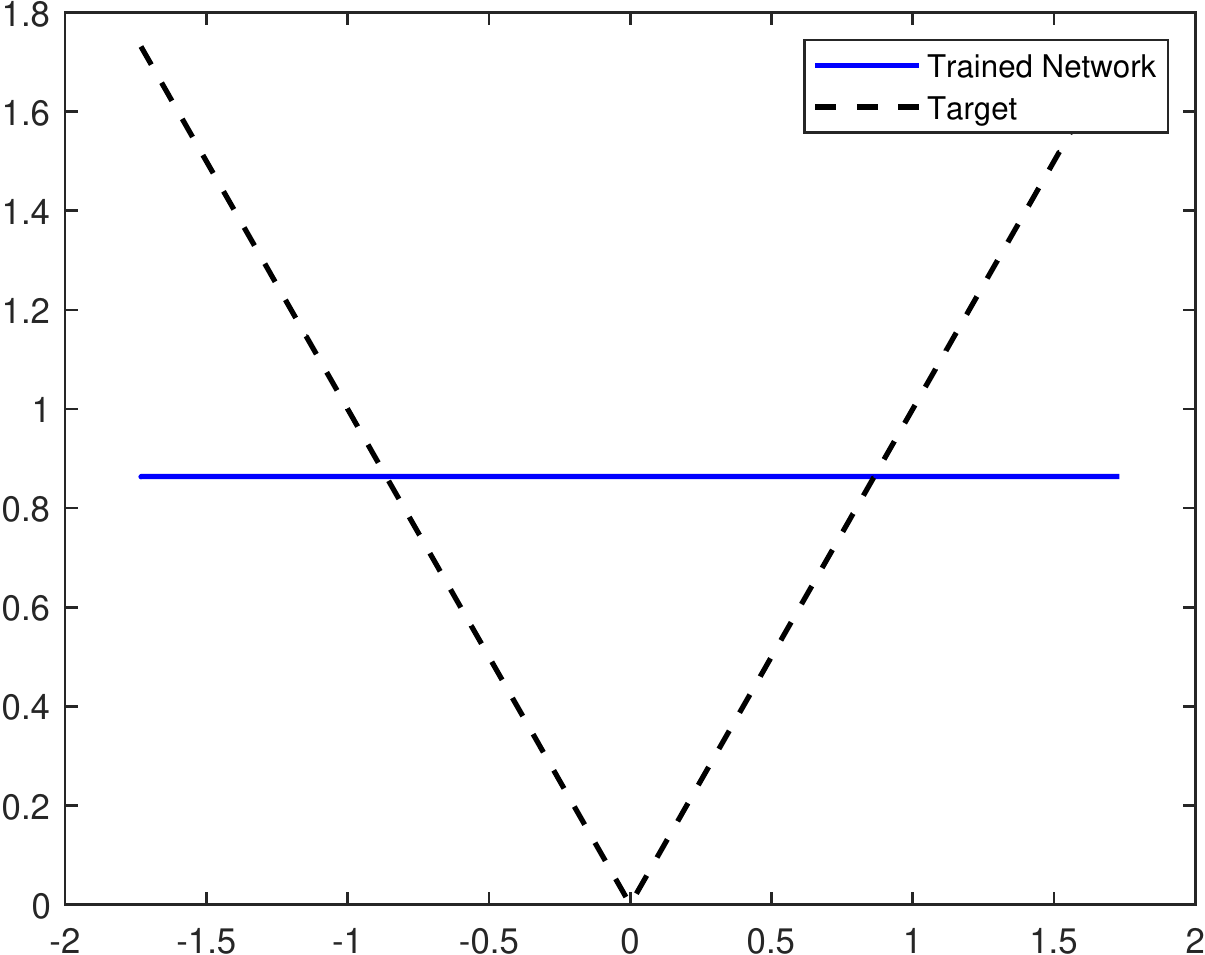}
	}
	\caption{(Left) The empirical probability that a network approximates $f_1(x)$ successfully and the probability that a network is trainable (Theorem~\ref{THM:MAIN})
	with respect to the size of width $n$.
	A trained network which falls in (middle) $\mathcal{F}_1$
	and (right) $\mathcal{F}_0$.}
	 \label{fig:absx-trained}
\end{figure}

In Figure~\ref{fig:absx-trained}, we show the approximation results for approximating $f_1(x)=|x|$.
On the left, we plot the empirical probability of successful training with respect to the value of width.
The empirical probabilities are obtained from 1,000 independent simulations
and a single simulation is regarded as a success if the test error is less than $10^{-2}$.
We also plot the trainability from Theorem~\ref{THM:MAIN}.
As expected, it provides an upper bound for the probability of successful training.
It is clear that 
the more the network is over-specified, 
the higher trainability is obtained.
Also, it can be seen that as the size of width grows, the empirical training success rate 
increases. This suggests that a successful training could be achieved (with high probability) by having a very high trainability. 
However, since it is only a necessary condition, 
although an initialized network is in $\mathcal{F}_j$ for $j \ge 2$, i.e., trainable, 
the final trained result could be in
either $\mathcal{F}_1$ or $\mathcal{F}_0$ as shown in
the middle and right of Figure~\ref{fig:absx-trained}, respectively.

\begin{figure}[!htbp]
	\centerline{
		\includegraphics[width=5.25cm]{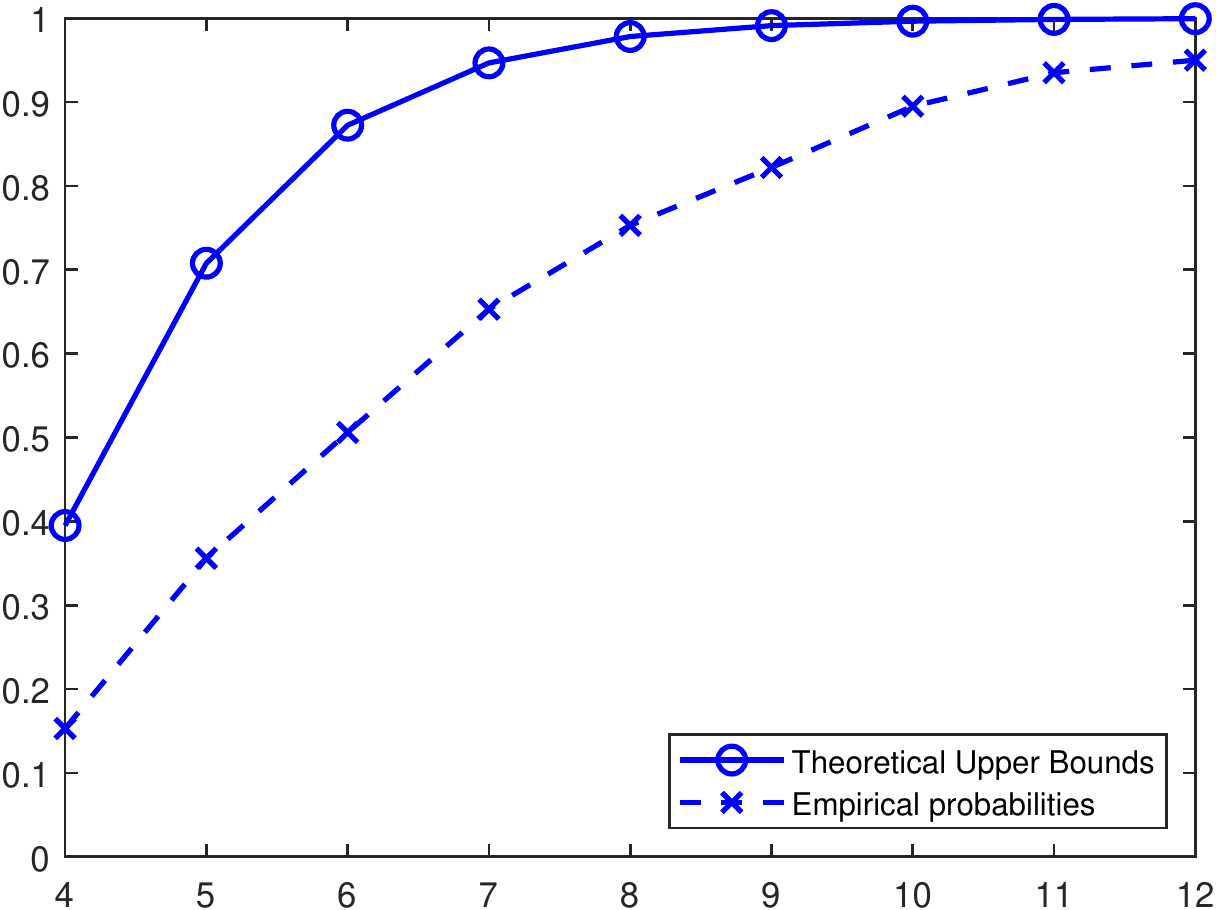}
		\includegraphics[width=4.95cm]{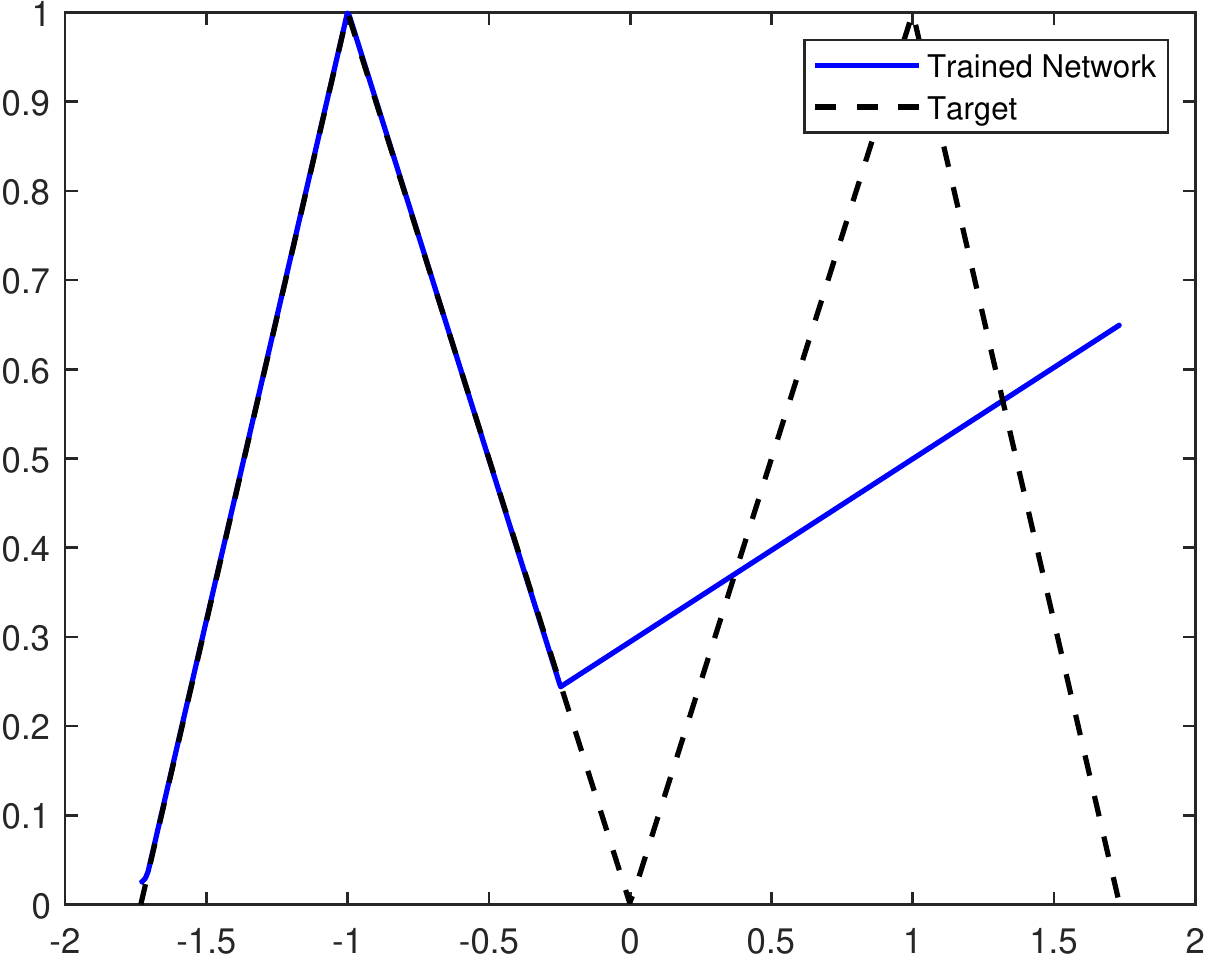}
		\includegraphics[width=4.95cm]{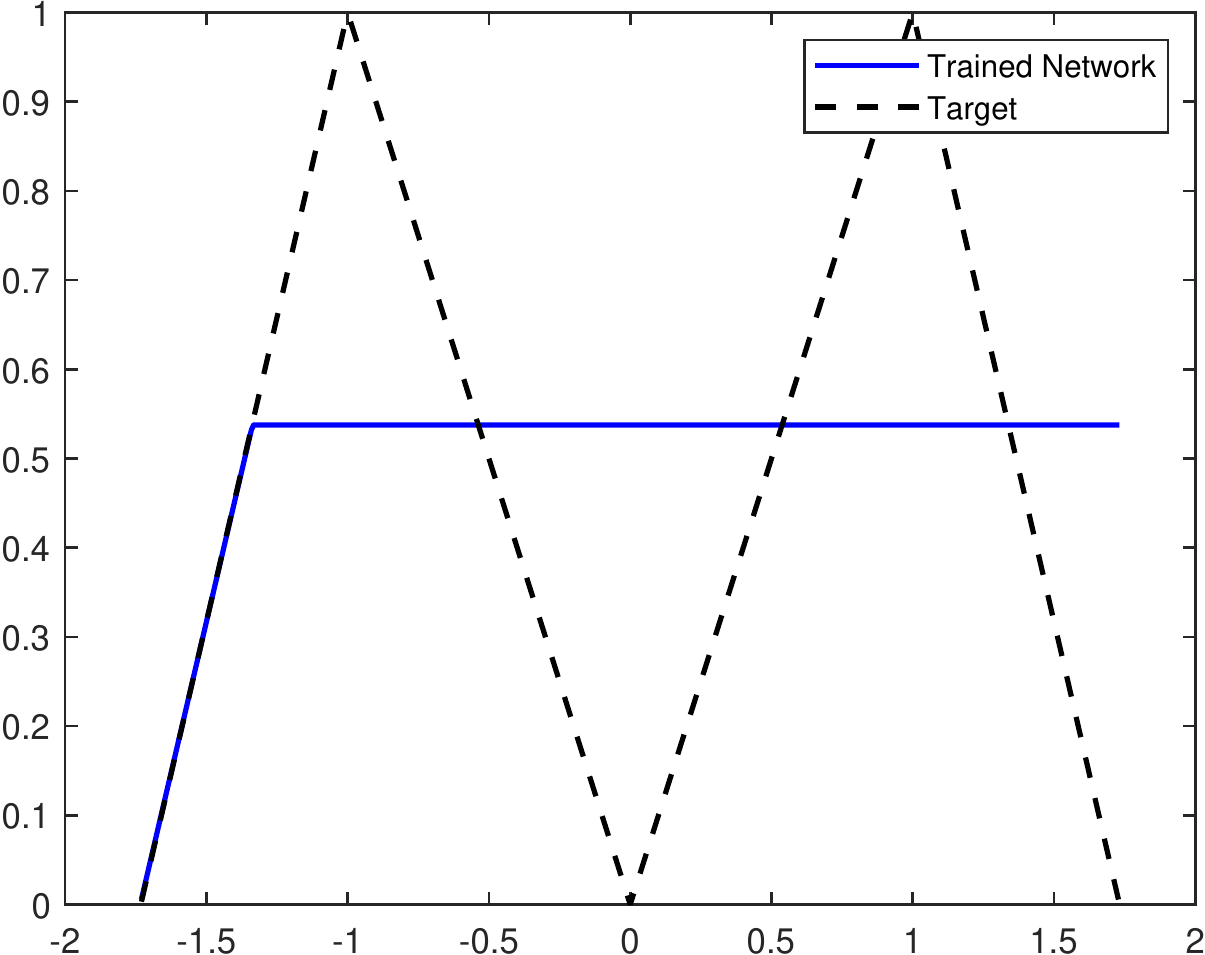}
	}
	\caption{(Left) The empirical probability that a network approximates $f_2(x)$ successfully 
	and the probability that a network is trainable (Theorem~\ref{THM:MAIN})
	with respect to the size of width $n$. 
	A trained network which falls in (middle) $\mathcal{F}_3$
	and (right) $\mathcal{F}_2$.
	}
	\label{fig:f2-local-min}
\end{figure}

Similar behavior is observed for approximating $f_2(x)$.
In Figure~\ref{fig:f2-local-min},
we show the approximation results for $f_2(x)$.
On the left, both the empirical probability of successful training and the trainability (Theorem~\ref{THM:MAIN}) are plotted with respect to the size of width.
Again, the trainability provides an upper bound for the probability of successful training.
Also, it can be seen that
the empirical training success rate 
increases, as the size of width grows.
On the middle and right,
we plot two of local minima
which a trainable network could end up with.
We remark that the choice of gradient-based optimization methods,
well-tuned learning rate, and/or other tunable optimization parameters could affect the empirical training success probability.
However, the maximum probability one can hope for is bounded by the trainability.
In all of our simulations, 
we did not tune any optimization hyper-parameters.

\subsection{Data-dependent Bias Initialization}
Next, we compare the training performance of three initialization methods. 
The first one is the `He initialization' \citep{he2015delving} without bias. This corresponds to 
$\bm{W}^1 \sim N(0,2/d_\text{in}), \bm{W}^2 \sim N(0,2/n), \bm{b}^1 = 0, \bm{b}^2 = 0$.
The second one is the `He initialization' with bias \eqref{initialization-He}.
This corresponds to 
$[\bm{W}^1,\bm{b}^1] \sim N(0,2/(d_\text{in}+1)), [\bm{W}^2, \bm{b}^2] \sim N(0,2/(n+1))$.
Here $n$ is the width of the 1st hidden layer.
The last one is the proposed data-dependent initialization described in the previous section. 
We use the parameters from \eqref{data-dependent-parameter}.
All results are generated under the same conditions except for the weights and biases initialization.

We consider the following $d_\text{in}=2$ test functions on $[-1,1]^2$: 
\begin{equation} \label{test-f}
\begin{split}
    f_3(\x) &= \sin(\pi x_1)\cos(\pi x_2)e^{-x_1^2-x_2^2}, \\
    f_4(\x) &= \sin(\pi(x_1-x_2))e^{x_1+x_2}.
\end{split}
\end{equation}
In all tests, we employ a shallow ReLU network of width 100 and it is trained over 25 randomly uniformly drawn points from $[-1,1]^2$.
We employ the gradient-descent method with moment with the square loss. 
The learning rate is a constant of $0.005$ and the momentum term is 0.9.  

\begin{figure}[!htbp] 
	\centerline{
		\includegraphics[width=7cm]{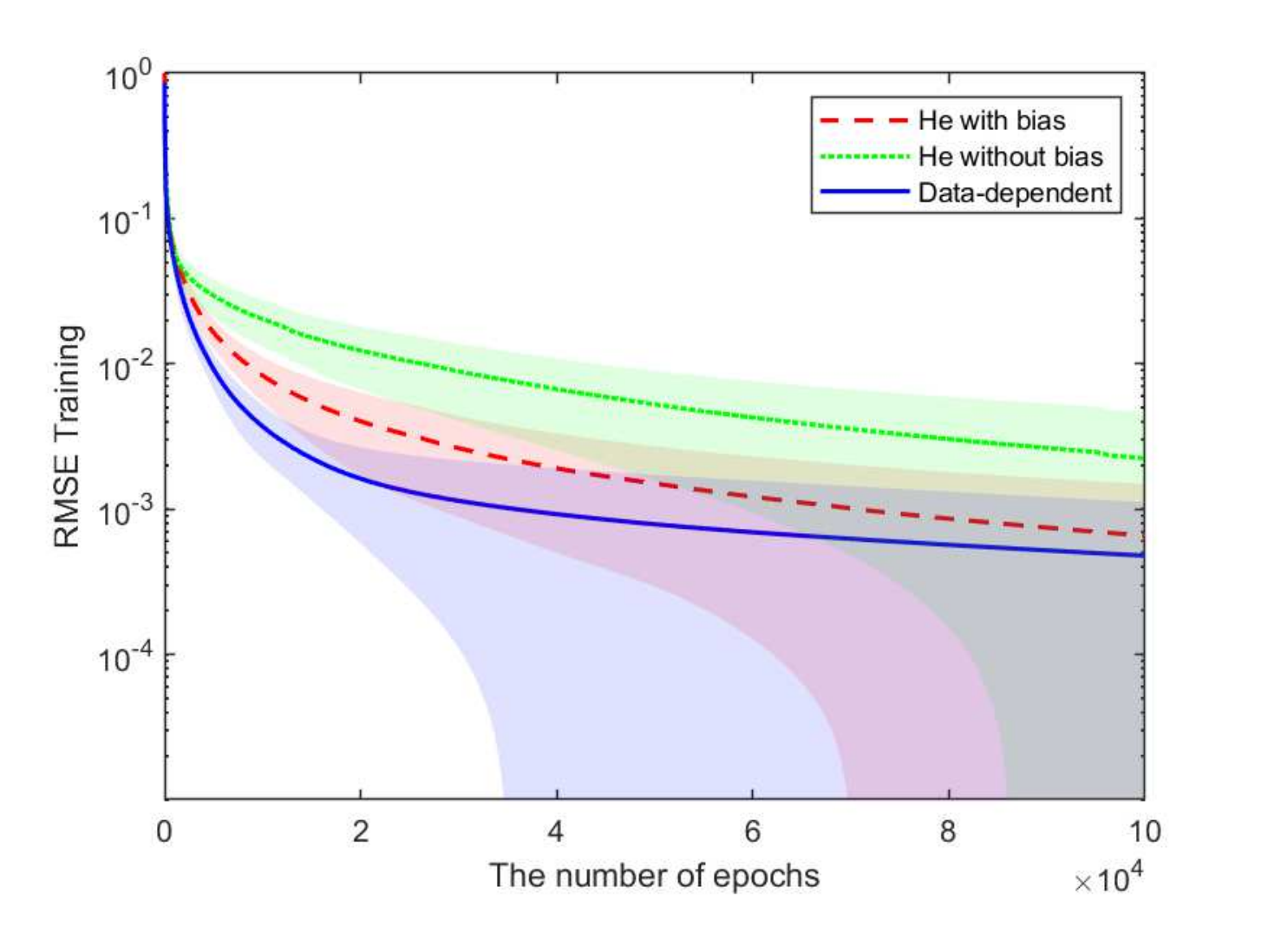}
		\includegraphics[width=7cm]{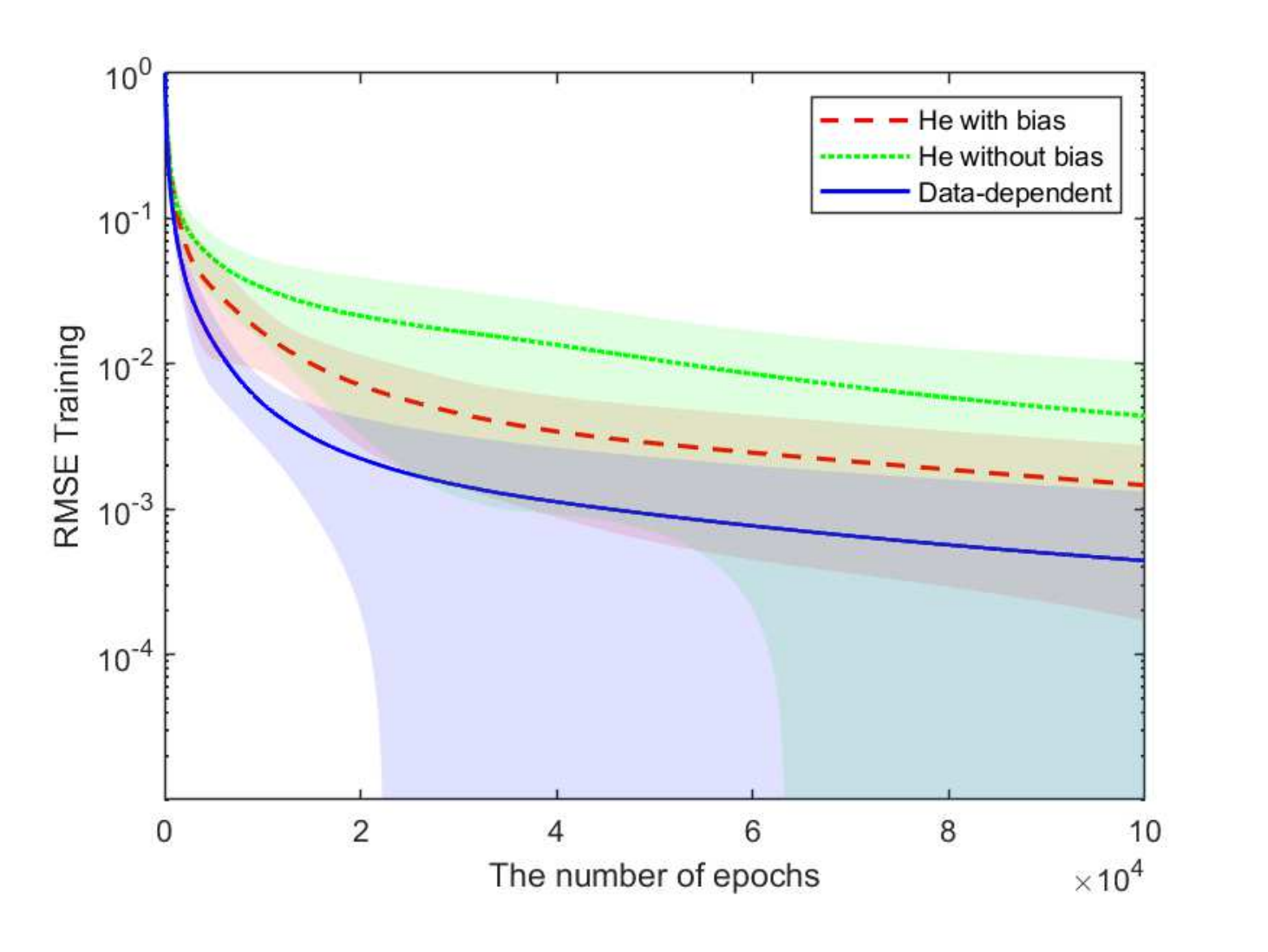}
	}
	\caption{The convergence of the root mean square error on the training data for approximating (left) $f_3$ and (right) $f_4$ with respect to the number of epochs of the gradient descent with moment
	by three different initialization methods.
	A shallow (1-hidden layer) ReLU network of width 100 is employed.
	The shaded area covers plus or minus one standard deviation from the mean.}
	\label{fig:2d-data-dependent}
\end{figure}

Figure~\ref{fig:2d-data-dependent} shows the mean of the RMSE on the training data from 10 independent simulations with respect to the number of epochs by three different initialization methods.
The shaded area covers plus or minus one standard deviation from the mean.
On the left and right, the results for approximating $f_3(x)$ and $f_4(x)$ are shown, respectively. 
We see that the data-dependent initialization not only results in the faster loss convergence but also achieves the smallest training loss. 
Also, the average number of dead neurons in the trained network is 
11 (He with bias), 0 (He without bias), and 
0 (Data-dependent) for $f_3$,
and 
12 (He with bias), 0 (He without bias), and 
0 (Data-dependent) for $f_4$.
Together with the example in section~\ref{sec:data-dependent}, 
all examples demonstrate the effectiveness of the proposed data-dependent initialization.


\section{Conclusion}
\label{sec:conclusion}
In this paper, we establish 
the trainability of ReLU neural networks,
a necessary condition for the successful training,
and propose a data-dependent initialization scheme for the better training.

Upon introducing two states of dead neurons; tentatively dead and permanently dead,
we define a trainable network.
A network is trainable if it has sufficiently small permanently dead neurons.
We show that a network being trainable is a necessary condition for the successful training.
We refer to the probability of a randomly initialized network being trainable as {\em trainability}.
The trainability serves as an upper bound of the probability of successful training.
We establish a general formulation for computing the trainability and derive the trainabilities of some special cases.
For shallow ReLU networks, by utilizing the computed trainability, 
we show that over-parameterization is both a neccessary and a sufficient condition for interpolating all training data, i.e.,
minimizing the loss.

Motivated by our theoretical results, we propose a data-dependent initialization scheme in the over-parameterized setting. 
The proposed method is designed to avoid both the dying ReLU neuron problem and 
to efficiently locate all neurons at the initialization for the faster training.
Numerical examples are provided to demonstrate the performance of our method.
We found that the data-dependent initialization method outperforms both the `He initialization' with and without bias in all of our tests.


\acks{This work is supported by the DOE PhILMs project (No.de-sc0019453), 
the AFOSR grant FA9550-17-1-0013, 
and the DARPA AIRA grant HR00111990025.}


\newpage

\appendix
\section{Proof of Lemma~\ref{THM:DEAD-STAY-DEAD}} \label{app:thm:dead-stay-dead}
\begin{proof}
	Suppose a ReLU neural network $\N(x)$ of width $N$ is initialized to be 
	$$
	\N(x;\theta) = \sum_{i=1}^n c_i\phi(\bm{w}_i^T\x + b_i) + \sum_{i=n+1}^N c_i\phi(\bm{w}_i^T\x + b_i) + c_0,
	$$
	where
	$\bm{\theta} = ((c_i,\bm{w}_i,b_i)_{i=1}^N, c_0)$ and
	the second term on the right is a constant function on $B_r(0)$. 
	Let $Z(x) = \sum_{i=n+1}^N c_i\phi(\bm{w}_i^T\x + b_i)$.
	Given a training data set $\mathcal{T}_m=\{(\x_i,y_i)\}_{i=1}^m$
	where $\{\x_i\}_{i=1}^m \subset B_r(0)$,
	and a loss metric $\ell:\mathbb{R}^{d_\text{out}}\times \mathbb{R}^{d_\text{out}} \mapsto \mathbb{R}$,
	the loss function is 
	$\mathcal{L}(\bm{\theta};\mathcal{T}_m) = \sum_{i=1}^m \ell(\N(\x_i;\bm{\theta}), y_i)$.
	The gradients of the loss function $\mathcal{L}$ with respect to parameters are
	\begin{equation}
	\frac{\partial}{\partial \theta}\mathcal{L}(\bm{\theta};\mathcal{T}_m)
	= \sum_{(x,y) \in \mathcal{T}_m} \ell'(\N(x;\bm{\theta}),y) \frac{\partial}{\partial \theta}\N(x;\bm{\theta}).
	\end{equation}
	Then for $i=1,\cdots, N$, we have
	$\frac{\partial}{\partial \bm{w}_i}\N(x;\bm{\theta}) = \phi'(\bm{w}_i^T\x + b_i) \x$
	and $\frac{\partial}{\partial b_i}\N(x;\bm{\theta}) = \phi'(\bm{w}_i^T\x + b_i)$.
	Since $Z(\x)$ is a constant function on $\Omega$, 
	for $i=n+1,\cdots,N$, we have $\phi'(\bm{w}_i^T\x + b_i) = 0$ for all $\x \in B_r(0)$.
	Therefore, any gradient-based optimization method does not update
	$(c_i,\bm{w}_i,b_i)_{i=n+1}^N$,
	which makes $Z(\x)$ remain a constant function in $B_r(0)$.
	
	It follows from Lemma 10 of \cite{lu2019dying}
	that with probability 1, a network is initialized to be a constant function
	if and only if there exists a hidden layer such that 
	all neurons are dead.
	Thus, all dead neurons cannot be revived through gradient-based training.
\end{proof}

\section{Proof of Lemma~\ref{THM:INTERPOLATION}} \label{app:THM:INTERPOLATION}
\begin{proof}
	Given a set of non-degenerate $m$ data, $\{\x_i,y_i\}_{i=1}^m$,
	for $d_{\text{in}} = 1$,
	suppose $x_1 < x_2 < \cdots < x_m$
	and for $d_{\text{in}} > 1$, 
	we choose a vector $\bm{w}$ such that 
	$\bm{w}^T\x_1 < \cdots < \bm{w}^T\x_m$.
	We note that one can always find such $\bm{w}$. 
	Let 
	$$
	S_{ij} = \{\bm{w} \in \mathbb{S}^{d_{\text{in}-1}} | \bm{w}^T(\x_i - \x_j) = 0 \}, \qquad i\ne j. 
	$$
	Since $\x_i$'s are distinct, 
	$S_{ij}$ is a Lebesgue measure zero set.
	Thus, $\cup_{1\le i < j \le m} S_{ij}$ is also a measure zero set.
	Therefore, 
	the Lebesgue measure of 
	$\cap_{1\le i < j \le m} S_{ij}^c$ 
	is positive and thus, it is nonempty. 
	Then, any vector $\bm{w} \in \cap_{1\le i < j \le m} S_{ij}^c$ 
	satisfies the condition.
	
	We recursively define shallow ReLU networks; for $j=0,\cdots,m$, 
	$$
	\N(\x;j) = y_1 + \sum_{i=1}^{j} c_i\phi(\bm{w}^T(\x - \x_i)),  \qquad c_i = \frac{y_{i+1} - \N(\x_{i+1};i-1)}{\bm{w}^T(\x_{i+1}-\x_i)}.
	$$
	Then it can be checked that 
	$\N(\x_j;m-1) = y_j$ for all $j$.
	Since $\bm{w}^T(x_j - x_k) \le 0$ for all $k \ge j$,
	$\N(\x_j;m-1) = \N(\x_j;j-1)$.
	Also, since $\N(\x_j;j-1) = \N(\x_j;j-2) + c_{j-1}\bm{w}^T(\x_j-\x_{j-1})$ 
	and $c_{j-1} = \frac{y_j - \N(\x_{j};j-2)}{\bm{w}^T(\x_j-\x_{j-1})}$,
	we have $\N(\x_j;m-1) = y_j$.
	
	Let $\{(\x_i,y_i)\}_{i=1}^{m}$ be the set of $(m+1)$ data such that 
	$\x_i = \alpha_i \x_1$ and $\alpha_i$'s are distinct.
	Also let 
	$\alpha_1 < \cdots < \alpha_m$ (after the reordering if necessary) and
	\begin{equation} \label{nd-data-cond}
	\frac{y_{i+2}-y_{i+1}}{\alpha_{i+1}-\alpha_{i}} \ne
	\frac{y_{i+1}-y_i}{\alpha_{i+1}-\alpha_i} \ne \frac{y_{i}-y_{i-1}}{\alpha_{i}-\alpha_{i-1}}, \forall i=2,\cdots,m-2.
	\end{equation}
	Suppose there exists a network $\N(\x; m-2)$ of width $(m-2)$ which interpolates all $m$ data.
	We note that a shallow ReLU network is a piece-wise linear function. 
	That is, whenever a slope in a direction needs to be changed, a new neuron has to be added.
	Since the number of neurons is $(m-2)$,
	the number of slope changes is at most $(m-2)$. 
	However, in order to interpolate the data set satisfying \eqref{nd-data-cond},
	the minimum number of slope changes 
	is $m-1$. 
	To be more precise,
	the network in the direction of $\x_1$ can be viewed as a one-dimensional network satisfying
	\begin{align*}
	\N(\x_i;m-2) = \N(\alpha_i\x_1;m-2) = y_i, \forall i.
	\end{align*}
	Since $\N(s\x_1;m-2)$ is a network of width $(m-2)$ in 1-dimensional input space (i.e., as a function of $s$)
	and it interpolates $m$ data satisfying \eqref{nd-data-cond},
	there must be at least $m-1$ slope changes in the interval $[\alpha_1,\alpha_m]$.
	However, since $\N$ has only $(m-2)$ width, this is impossible.
	Therefore, any shallow ReLU network of width less than $(m-2)$ cannot interpolate $m$ data points which satisfy \eqref{nd-data-cond}.
\end{proof}

\section{Proof of Theorem~\ref{THM:OVERPARA-IFF}}
\label{app:THM:OVERPARA-IFF}
\begin{proof}
	It had been shown in several existing works \citep{du2018gradient-shallow, oymak2019towards,li2018learning}
	that with probability at least $1-\delta$ over the initialization, an over-parameterized shallow ReLU network
	can interpolate all training data
	by the (stochastic) gradient-descent method.
	In other words,
	over-parameterization is a sufficient condition for interpolating all training data with probability at least $1-\delta$.
	
	By Lemma~\ref{THM:INTERPOLATION},
	in order to interpolate $(m+1)$ data points,
	a shallow ReLU network having at least width $m$ is required.
	However, the probability that 
	an initialized ReLU network of width $m$ has $m$ active neurons is 
	$$
	\text{Pr}(\m_1 = m) = (1-\hat{p}_{d_\text{in}}(r))^m,
	$$
	which decays exponentially in $m$.
	It follows from Lemma~\ref{LEM:DYINGPROB} that 
	$\hat{p}_{d_\text{in}}(r) > \frac{(\sin\alpha_r)^{d_\text{in}}}{\pi d_\text{in}}$
	where $\alpha_r = \tan^{-1}(1/r)$.
	From the assumption of \eqref{eqn:overpara-iff}, we have
	$$
	\text{Pr}(\m_1 = m) = (1-\hat{p}_{d_\text{in}}(r))^m 
	< \left(1 - \frac{(\sin\alpha_r)^{d_\text{in}}}{\pi d_\text{in}}\right)^m < 1-\delta.
	$$
	That is, the trainability is less than $1-\delta$. 
	Therefore, 
	over-parameterization is 
	required to guarantee, with probability at least $1-\delta$,
	that 
	at least $m$ neurons are active 
	at the initialization. 
	Therefore, 
	over-parameterization is a necessary
	condition for interpolating all training data.
\end{proof}

\section{Proof of Theorem~\ref{THM:MAIN}} \label{app:THM:MAIN}
\begin{proof}
    It follows from Lemma~\ref{def:ANPDist}, 
    since $\pi_0 = [0,\cdots,0,1]$, it suffices to compute the last row of the stochastic matrix $\text{P}_1$. 
    For completeness, we set 
    $(\text{P}_1)_{i,:}=[1,0,\cdots,0]$
    for $i=1,\cdots,n_0$.
    
    Suppose the `He initialization' without bias is used. 
    Since $\x \in B_r(0)$, for any $\bm{w}$,
    there exists some $\x \in B_r(0)$ such that $\bm{w}^T\x > 0$. 
    Therefore, no ReLU neuron will be born dead. Hence, 
    $(\text{P}_1)_{n_0+1,:} = [0,\cdots,1]$.
    
    Suppose the `He initialization' with bias is used.
    Since each hidden neuron is independent, 
    $\text{m}_1$ follows a binomial distribution $B(n_1,p)$.
    Here $p$ represents the born dead probability of a single ReLU neuron in the 1st hidden layer. 
    By Lemma~\ref{LEM:DYINGPROB}, 
    $p = \hat{p}_{n_0}(r)$ and this completes the proof.
\end{proof}
The proof is completed by the following lemma.
\begin{lemma} \label{LEM:DYINGPROB}
	Suppose all training data inputs are from $B_r(0) = \{x \in \mathbb{R}^d | \|x\| \le r \}$ 
	and the weights and the biases are independently initialized 
	from a zero mean normal distribution $N(0,\sigma^2)$.
	Then, the probability that a single ReLU neuron dies at the initialization is 
	\begin{equation}
	\hat{p}_d
	= \frac{1}{\sqrt{\pi}}\frac{\Gamma((d+1)/2)}{\Gamma(d/2)}\int_0^{\alpha_r} (\sin u)^{d-1} du < \frac{1}{2},
	\end{equation}
	where $\alpha_r = \tan^{-1}(r^{-1})$.
	Furthermore,
	\begin{equation}
	\frac{1}{\pi d}(\sin \alpha)^d
	\le \hat{p}_d \le \sqrt{\frac{d}{2\pi}}\alpha (\sin \alpha)^{d-1}.
	\end{equation}
\end{lemma}
\begin{proof}[Proof of Lemma~\ref{LEM:DYINGPROB}]
    Let $\phi(\bm{w}\x+b)$ be a single ReLU neuron 
    where $\bm{w}_i, b \sim N(0,\sigma^2)$.
    Note that in order for a single ReLU neuron to die in $B_r(0)$, 
    for all $\x \in B_r(0)$, $\bm{w}\x + b < 0$.
    Therefore, it suffices to calculate 
    $$
    \hat{p}_{d_\text{in}} = \text{Pr}(\bm{w}\x + b < 0, \forall \x \in B_r(0)).
    $$
    Let $\bm{v} = [\bm{w}, b] \in \mathbb{R}^{d_\text{in}+1}$.
    Since $\bm{w}_i$'s and $b$ are iid normal, 
    $\bm{s}:=\bm{v}/\|\bm{v}\|$
    follows the uniform distribution on the unit hypersphere $\mathbb{S}^{d_\text{in}}$, i.e., $\bm{s} \sim \mathcal{U}(\mathbb{S}^{d_\text{in}})$.
    Also, since $\bm{s} \overset{d}{=} -\bm{s}$, we have
    $$
    \hat{p}_{d_\text{in}} = \text{Pr}(\langle \bm{s}, [\x, 1] \rangle < 0, \forall \x \in B_r(0)) = \text{Pr}(\langle \bm{s}, [\x, 1] \rangle > 0, \forall \x \in B_r(0)).
    $$
    Let 
    \begin{equation}
        \mathcal{A} = \left\{\bm{s} \in \mathbb{S}^{d_\text{in}} \bigg| \langle \bm{s}, \bm{v} \rangle > 0, \forall \bm{v} \in B_r(0) \times \{1\}  \right\}.
    \end{equation}
    Then $\hat{p}_{d_\text{in}} = \text{Pr}(\mathcal{A})$.
    Let $\bm{s} \in \mathbb{S}^{d_\text{in}}$. 
    If $\bm{s}_{d_\text{in}+1} \le 0$, then $\bm{s} \not\in \mathcal{A}$. This is because there exists $\bm{v} = (0,\cdots,0,1) \in B_r(0)\times \{1\}$ 
    such that $\langle \bm{s}, \bm{v} \rangle \le 0$. 
    Suppose $\bm{s}_{d_\text{in}+1} > 0$ and let $r_{\bm{s}} = 1/\bm{s}_{d_\text{in}+1}$. 
    Then 
    $\tilde{\bm{s}}:=\frac{1}{\bm{s}_{d_\text{in}+1}}\bm{s} \in B_{r_{\bm{s}}}(0) \times \{1\}$.
    
    We can express any $\x \in B_r(0)$ in the spherical coordinate system, i.e.,
    \begin{align*}
        \x_1 &= t\cos(\theta_1), \\
        \x_2 &= t\sin(\theta_1)\cos(\theta_2), \\ 
        &\vdots \\
        \x_{d_\text{in}-1} &= t\sin(\theta_1)\cdots \sin(\theta_{d_\text{in}-2})\cos(\theta_{d_\text{in}-1}), \\
        \x_{d_\text{in}} &= t\sin(\theta_1)\cdots \sin(\theta_{d_\text{in}-2})\sin(\theta_{d_\text{in}-1}).
    \end{align*}
    Since $\bm{s}$ is a uniform random variable from $\mathbb{S}^{d_\text{in}}$, it is coordinate-free. 
        Thus, let $\tilde{\bm{s}} = (r_{\bm{s}},0,\cdots,0, 1)$ for some  $0 \le r_{\bm{s}}$
    and $\bm{v} =[\x, 1] \in B_r(0)\times \{1\}$.
    Then
    \begin{align*}
        \langle \tilde{\bm{s}}, \bm{v} \rangle = 1 + r_{\bm{s}}t\cos(\theta_1), \qquad 0 \le t \le r, 0 \le \theta_1 \le 2\pi.
    \end{align*}
    In order for $\bm{s} \in \mathcal{A}$, 
    $r_{\bm{s}} < \frac{1}{r}$ has to be satisfied. 
    Therefore, 
    $$
    \mathcal{A} = \left\{ \frac{\tilde{\bm{s}}}{\|\tilde{\bm{s}}\|} \in \mathbb{S}^{d_\text{in}} \big| \tilde{\bm{s}} \in B_{1/r}(0)\times \{1\}\right\}. $$
    Let $\text{Surf}(\mathbb{S}^d)$ be the surface area of $\mathbb{S}^d$. It is known that 
    $$
    \text{Surf}(\mathbb{S}^d) = \frac{2\pi^{\frac{d+1}{2}}}{\Gamma(\frac{d+1}{2})},
    $$
    where $\Gamma$ is the gamma function.
    Then
    \begin{align*}
        \hat{p}_{d_\text{in}} &= \text{Pr}(\mathcal{A}) \\
        &= \frac{1}{\text{Surf}(\mathbb{S}^{d_\text{in}})}
        \int_{\mathcal{A}} d^{d_\text{in}+1}\mathbf{S} \\
        &=\frac{1}{\text{Surf}(\mathbb{S}^{d_\text{in}})}
        \int_{\theta_{d_\text{in}} = 0}^{2\pi}
        \int_{\theta_{d_\text{in}-1} = 0}^{\pi}
        \cdots 
        \int_{\theta_{2} = 0}^{\pi}
        \int_{\theta_1 = 0 }^\alpha 
        d^{d_\text{in}+1}\mathbf{S} \\
        &=\frac{\text{Surf}(\mathbb{S}^{d_\text{in}-1})}{\text{Surf}(\mathbb{S}^{d_\text{in}})} \int_{0}^\alpha \sin^{d-1}\theta d\theta,
    \end{align*}
    where $\alpha = \tan^{-1}(1/r)$ and
    $$
    d^{d_\text{in}+1}\mathbf{S} = \sin^{d_\text{in}-1}\theta_1 \sin^{d_\text{in}-2}\theta_2 \cdots \sin \theta_{d_\text{in}-1} d\theta_1 d\theta_2 \cdots d\theta_{d_\text{in}}.
    $$
    Note that $g(d) = \frac{\text{Surf}(\mathbb{S}^{d-1})}{\text{Surf}(\mathbb{S}^{d})}$ is
    bounded above by $\sqrt{\frac{d}{2\pi}}$ for all $d \ge 1$ \citep{leopardi2007distributing}
    and $g(d)$ is monotonically increasing.
    Thus, we have an upper bound of $\hat{p}_d$ as
    $$
    \hat{p}_d \le \sqrt{\frac{d}{2\pi}}\alpha (\sin \alpha)^{d-1}.
    $$
    For a lower bound, 
    it can be shown that for any $\theta \in [0,\pi]$,
    $$
    \int_0^\theta (\sin x)^{d-1} dx \ge \frac{1}{d}(\sin \theta)^d.
    $$
    Thus, we have 
    $$
    \hat{p}_d \ge g(d)\frac{(\sin \alpha)^d}{d} \ge g(1)\frac{(\sin \alpha)^d}{d} = \frac{(\sin \alpha)^d}{\pi d},
    $$
    which completes the proof.
\end{proof}

\section{Probability Distribution of the Number of Active ReLU Neurons}  \label{sec:analysis}
In order for calculating the trainability, 
we first present the results for the distribution of the number of active neurons.
Understanding how many neurons will be active at the initialization is not only directly related to the trainability of a ReLU network, 
but also suggests 
how much over-specification or over-parameterization shall be needed for training. 
Given a $L$-layer ReLU network with $\bm{n}=(n_0,n_1,\cdots,n_L)$ architecture,
let $\mathfrak{m}_t$ be the number of active neurons at the $t$-th hidden layer
and $\pi_t$ be its probability distribution. 


Then, the distribution of $\mathfrak{m}_t$ can be identified as follows.
\begin{lemma} \label{def:ANPDist}
	Let $\bm{V}^t = [\bm{W}^t, \bm{b}^t]$ be the parameter (weight and bias) matrix in the $t$-th layer.
	Suppose $\{\bm{V}^t\}_{t=1}^L$ is randomly independently initialized 
	and each row of $\bm{V}^t$ is independent of any other row and follows an identical distribution.  
	Then, 
	the probability distribution of 
	the number of active neurons $\m_j$ at the $j$-th hidden layer can be expressed as 
	\begin{equation} 
	\pi_{j} = \pi_0\text{P}_1\text{P}_2\cdots \text{P}_{j}, \qquad
	(\pi_j)_i = \text{Pr}(\mathfrak{m}_j = i),
	\end{equation}
	where $\pi_0 =[0,\cdots,0,1]$, $\text{P}_{t}$ is the stochastic matrix of size $(n_{t-1}+1)\times (n_t+1)$ whose $(i+1,j+1)$-entry is 
	$\text{Pr}(\mathfrak{m}_t = j | \mathfrak{m}_{t-1} = i)$.
	Furthermore, the stochastic matrix $P_t$ is expressed as 
	\begin{equation} \label{def:stochasticMAT}
	(\text{P}_t)_{(i+1,j+1)}= \binom{n_t}{j} \mathbb{E}_{t-1}\left[(1-\mathfrak{p}_t({A}_{t-1}^i))^j\mathfrak{p}_t({A}_{t-1}^i)^{n_t-j}\right],
	\end{equation}
	where 
	$\mathbb{E}_{t-1}$ is the expectation with respect to $\{\bm{W}^i, \bm{b}^i\}_{i=1}^{t-1}$ and 
	$\mathfrak{p}_t({A}_{t-1}^i)$ is the conditional born dead probability (BDP) of a neuron in the $t$-th layer given 
	the event where exactly $i$ neurons are active in the $(t-1)$-th layer.
\end{lemma}
\begin{proof}[Proof of Lemma~\ref{def:ANPDist}]
	By the law of total probability, it readily follows that for $j=0,\cdots,n_t$,
	$$
	\text{Pr}(\mathfrak{m}_t = j) = \sum_{k=0}^{n_{t-1}} \text{Pr}(\mathfrak{m}_t = j | \mathfrak{m}_{t-1}=k)\text{Pr}(\mathfrak{m}_{t-1} = k),
	$$
	which gives $\pi_t = \pi_{t-1}\text{P}_{t}$.
	By recursively applying it, we obtain $\pi_t = \pi_0\text{P}_1\cdots \text{P}_t$.
	
	For each $t$ and $i$, 
	let $A_{t-1}^i$ be the event where exactly $i$ neurons are active in the $(t-1)$-th layer
	and $D_t^k$ be the event where the $k$-th neuron in the $t$-th layer is dead. 
	Since each row of $\bm{V}^t$ is iid and $\{\bm{V}^t\}_{t=1}^L$ is independent,
	$\text{Pr}(D_t^k|A_{t-1}^i) = \text{Pr}(D_t^j|A_{t-1}^i)$ for any $k, j$.
	We denote the conditional BDP of a neuron in the $t$-th layer given ${A}_{t-1}^i$
	as $\mathfrak{p}_t({A}_{t-1}^i)$.
	From the independent row assumption, the stochastic matrix $P_t$ can be expressed as 
	\begin{equation*}
	\text{Pr}(\mathfrak{m}_t = j | \mathfrak{m}_{t-1} = i)
	= (\text{P}_t)_{(i+1,j+1)}= \binom{n_t}{j} \mathbb{E}_{t-1}\left[(1-\mathfrak{p}_t({A}_{t-1}^i))^j\mathfrak{p}_t({A}_{t-1}^i)^{n_t-j}\right],
	\end{equation*}
	where $\mathbb{E}_{t-1}$ is the expectation with respect to $\{\bm{W}^i, \bm{b}^i\}_{i=1}^{t-1}$.
\end{proof}

Lemma~\ref{def:stochasticMAT} indicates that $\mathfrak{p}_t({A}_{t-1}^i)$ is a fundamental quantity for the complete understanding of $\pi_j$.

As a first step towards understanding $\pi_j$, we calculate the exact probability distribution $\pi_1$ of the number of active neurons in the 1st hidden layer. 
\begin{lemma} \label{THM:PI_1}
	Given a ReLU network having 
	$\bm{n}=(n_0,n_1,\cdots,n_L)$ architecture, suppose the training input domain is $B_r(0)$. 
	If either the `normal' \eqref{initialization-He} or 
	the `unit hypersphere' \eqref{initialization-nsphere}
	initialization without bias is used in the 1st hidden layer, we have
	$$
	(\pi_1)_j = \text{Pr}(\mathfrak{m}_1=j) = \delta_{j,n_1}.
	$$
	If either the `normal' \eqref{initialization-He} or 
	the `unit hypersphere' \eqref{initialization-nsphere} with bias
	is used in the 1st hidden layer, $\mathfrak{m}_1$ follows a binomial distribution with parameters $n_1$ and  $1-\hat{p}_{n_0}(r)$,
	where
	\begin{equation}
	\hat{p}_d(r)
	= \frac{1}{\sqrt{\pi}}\frac{\Gamma((d+1)/2)}{\Gamma(d/2)}\int_0^{\alpha_r} (\sin \theta)^{d-1} d\theta, \qquad
	\alpha_r = \tan^{-1}(r^{-1}),
	\end{equation}
	and $\Gamma(x)$ is the Gamma function.
\end{lemma}
\begin{proof}
	The proof readily follows from Lemma~\ref{LEM:DYINGPROB}. 
\end{proof}
We now calculate $\pi_2$ for a ReLU network at $d_\text{in}=1$. 
Since the bias in each layer can be initialized in different ways, 
we consider some combinations of them.
\begin{lemma} \label{THM:PI_2}
	Given a ReLU network having 
	$\bm{n}=(1,n_1,n_2,\cdots,n_L)$ architecture, suppose the training input domain is $B_r(0)$. 
	\begin{itemize}[leftmargin=*]
		\item Suppose the `unit hypersphere' \eqref{initialization-nsphere} initialization without bias is used in the 1st hidden layer.
		\begin{enumerate}
			\item If the `normal' \eqref{initialization-He} initialization without bias is used in the 2nd hidden layer, 
			the stochastic matrix $\text{P}_2$ is
			$(\text{P}_2)_{i,:}=[1,0,\cdots,0]$ for $1\le i \le n_1$ and
			$$
			(\text{P}_2)_{n_1+1,j+1}=
			\binom{n_2}{j}\left[\left(1-\frac{1}{2^{n_1-1}}\right)\frac{3^j}{4^{n_2}} + \frac{1}{2^{n_1+n_2-1}}\right], 0\le j \le n_2.
			$$
			\item If the `normal' \eqref{initialization-He} initialization with bias is used in the 2nd hidden layers, the stochastic matrix $\text{P}_2$ is
			$(\text{P}_2)_{i,:}=[1,0,\cdots,0]$ for $1\le i \le n_1$ and
			$$
			(\text{P}_2)_{n_1+1,j+1}=\binom{n_2}{j}\mathbb{E}_{s}\left[(1-\mathfrak{p}_2(s))^j\mathfrak{p}_2(s)^{n_2-j}\right], \quad 0 \le j \le n_2, 
			$$
			where $s \sim B(n_1,1/2)$, $\alpha_s = \tan^{-1}(\frac{s}{n_1-s})$, $g(x) = \sin(\tan^{-1}(x))$, and
			\begin{align*}
			\mathfrak{p}_2(s) = \frac{1}{2} + \left[\int_{\frac{\pi}{2}}^{\pi+\alpha_{s}} \frac{g(r\sqrt{s}\cos(\theta))}{4\pi}d\theta + \int_{\pi+\alpha_{s}}^{2\pi} \frac{g(r\sqrt{n_1-s}\sin(\theta))}{4\pi}d\theta \right].
			\end{align*}
		\end{enumerate}
		\item Suppose the `unit hypersphere' \eqref{initialization-nsphere} initialization with bias is used in the 1st hidden layer. 
		\begin{enumerate}
			\item If the `normal' \eqref{initialization-He} initialization without bias is used in the 2nd hidden layer and $n_1=1$,
			the stochastic matrix $\text{P}_2$ is
			$(\text{P}_2)_{1,:}=[1,0,\cdots,0]$ and
			$$
			(\text{P}_2)_{2,:}=\text{Binomial}(n_2,1/2).
			$$
			\item If the `normal' \eqref{initialization-He} initialization with bias is used in the 2nd hidden layers and $n_1=1$, the stochastic matrix $\text{P}_2$ is
			$(\text{P}_2)_{1,:}=[1,0,\cdots,0]$
			and
			\begin{equation*}
			(\text{P}_2)_{2,j+1} = \binom{n_2}{j}\mathbb{E}_{\omega}\left[(1-\mathfrak{p}_2(\omega))^j\mathfrak{p}_2(\omega)^{n_2-j}\right], \quad 0 \le j \le n_2,   
			\end{equation*}
			where
			$\alpha_r = \tan^{-1}(r)$, $\omega \sim \text{Unif}\left(0,\frac{\pi}{2}+\alpha_r\right)$,
			$g(x) = \tan^{-1}(\frac{1}{\sqrt{r^2+1}\cos(x)})$,
			and
			\begin{align*}
			\mathfrak{p}_2(\omega) =
			\begin{cases}
			\frac{1}{4} + \frac{g(\omega - \alpha_r)}{2\pi}, 
			&\text{if } 
			\omega \in \left[\frac{\pi}{2}-\alpha_r, \frac{\pi}{2}+\alpha_r\right), \\
			\frac{1}{4} +\frac{g(\omega - \alpha_r) + \tan^{-1}\left(\sqrt{r^2+1}\cos(\omega + \alpha_r)\right)}{2\pi},
			& \text{if }
			\omega \in \left[0, \frac{\pi}{2}-\alpha_r\right).
			\end{cases}
			\end{align*}
		\end{enumerate}
	\end{itemize}
	Then $\pi_2 = \pi_1\text{P}_2$ where $\pi_1$ is defined in Lemma~\ref{THM:PI_1}.
\end{lemma}
\begin{proof}[Proof of Lemma~\ref{THM:PI_2}]
    Since $\pi_1$ is completely characterized in 
    Lemma~\ref{THM:PI_1},
    it suffices to calculate the stochastic matrix $\text{P}_2$, as $\pi_2 = \pi_1\text{P}_2$.
    From Equation~\ref{def:stochasticMAT},
    it suffices to calculate 
    the BDP $\mathfrak{p}_2(A_1^i)$ of a ReLU neuron at the 2nd layer given $A_1^i$.
    
    We note that 
    if $\bm{z} = [x,1]$ where $x \in B_r(0)=[-r,r]$
    and $\bm{v} = [w,b] \sim \mathcal{U}(\mathbb{S}^{1})$,
    then
    \begin{equation} \label{eqn-pd1}
        \begin{split}
            P(\phi(\bm{v}^T\bm{z}) = \bm{v}^T\bm{z}, \forall \bm{z} \in B_r(0)\times \{1\}) &= \hat{p}_1(r) = \frac{\tan^{-1}(1/r)}{\pi}, \\
    P(\phi(\bm{v}^T\bm{z}) = \bm{v}^T\bm{z}\mathbb{I}_{x \in [a,b]}, \forall \bm{z} \in B_r(0)\times \{1\})  
    &= \frac{1}{4} + \frac{\tan^{-1}(1/b) + \tan^{-1}(a)}{2\pi}.
        \end{split}
    \end{equation}

    First, let us consider the case where  
    the `unit hypersphere' initialization without bias is used for the 1st hidden layer. 
    Note that since $x \in [-r,r]$, i.e., $d_\text{in}=1$, we have
    $A_1^j = \emptyset$ for $0 \le j < n_1$
    and $A_1^{n_1} = \{1,-1\}^{n_1}$.
    Also, note that if ${w}_{j}$'s are iid normal, 
    $\sum_{j=1}^s w_{j} \overset{d}{=} \sqrt{s}w$
    where $w \overset{d}{=} {w}_{1}$.
    For fixed $\omega \in A_1^{n_1}$, 
    a single neuron in the 2nd layer is
    \begin{equation}\label{app:THM:PI_2:eqn-NB}
        \phi\left(\sum_{j=1}^{s}\bm{w}^2_{1,j}\phi(x) + \sum_{j=s+1}^{n_1}\bm{w}^2_{1,j}\phi(-x) + b\right)
    \overset{d}{=}
    \phi\left(\sqrt{s}{w}_{1}\phi(x) + \sqrt{n_1-s}w_{2}\phi(-x) + b\right),
    \end{equation} 
    where $s$ is the number of 1's in $\omega$. 
    If the `normal' initialization without bias is used for the 2nd hidden layer,
    we have 
    \begin{equation*}
        \mathfrak{p}_2(\omega) = \begin{cases}
    \frac{1}{2}, & \text{if } \omega = \pm[1,\cdots,1]^T, \\
    \frac{1}{4}, & \text{otherwise}.
    \end{cases} 
    \end{equation*}
    Also, $s \sim \text{Binomial}(n_1,1/2)$.
    Thus, for $j=0,\cdots,n_2$,
    \begin{align*}
        \text{Pr}(\text{m}_2=j|\text{m}_1=n_1) &=
    \binom{n_2}{j}\mathbb{E}_1 \left[(1-\mathfrak{p}_2(A_1^{n_1}))^j(\mathfrak{p}_2(A_1^{n_1}))^{n_2-j} \right] \\
    &=
    \binom{n_2}{j}\mathbb{E}_s\left[(1-\mathfrak{p}_2(A_1^{n_1}))^j(\mathfrak{p}_2(A_1^{n_1}))^{n_2-j} \right] \\
    &= \binom{n_2}{j}\left[\frac{1}{2^{n_1-1}}\frac{1}{2^{n_2}} + \left(1-\frac{1}{2^{n_1-1}}\right)\frac{3^j}{4^{n_2}}\right].
    \end{align*}
    Suppose the `normal' initialization with bias is used for the 2nd hidden layer.
    It follows from \eqref{app:THM:PI_2:eqn-NB} that
    \begin{equation*}
        \mathfrak{p}_2(\omega) = \text{Pr}(w_1\sqrt{s}\phi(x) + w_2\sqrt{n_1-s}\phi(-x) + b < 0, \forall x \in [-r,r] | \bm{W}^1 \text{ has $s$ 1's.}).
    \end{equation*}
    Let $\bm{z} = (\sqrt{s}\phi(x), \sqrt{n_1-s}\phi(-x),1)$
    and $\bm{v} = (w_1, w_2, b)$. 
    Without loss of generality, we normalize $\bm{v}$.
    Then $\bm{v} \sim \mathbb{S}^2$ and we write it as
    $$
    \bm{v}=(\cos \theta \sin\alpha, \sin \theta \sin\alpha, \cos\alpha),
    $$ where $\theta \in [0,2\pi]$ and $\alpha \in [0,\pi]$. 
    Since $\bm{v} \overset{d}{=} -\bm{v}$, it suffices to compute 
    $$
    \text{Pr}(\bm{v}^T\bm{z} > 0, \forall \bm{z} | \bm{W}^1 \text{ has $s$ 1's}).
    $$
    Also, note that
    \begin{align*}
        \bm{v}^T\bm{z} &= \begin{cases}
        \sqrt{s}\phi(x)\cos\theta \sin\alpha + \cos \alpha, & \text{if } x > 0, \\
        \sqrt{n_1-s}\phi(-x)\sin\theta \sin\alpha + \cos \alpha, 
        & \text{if } x < 0,
        \end{cases} \\
        &= \begin{cases}
        \sqrt{1 + sx^2\cos^2 \theta}\cos(\alpha - \beta), & \text{if } x > 0, \\
        \sqrt{1 + (n_1-s)x^2\sin^2 \theta}\cos(\alpha - \beta), 
        & \text{if } x < 0,
        \end{cases}
    \end{align*}
    where $\tan \beta = \sqrt{s}x\cos \theta$ if $x>0$
    and $\tan \beta = \sqrt{n_1-s}\sin \theta$ if $x<0$.
    Given $\bm{W}^1$ which has $s$ 1's, 
    the regime in $\mathbb{S}^2$, where $\bm{v}^T\bm{z} > 0$ for all $\bm{z}$, is
    \begin{align*}
        \text{For } \omega &\in [0, \frac{\pi}{2}], \alpha \in [0, \frac{\pi}{2}], \\
        \text{For } \omega &\in [\frac{\pi}{2}, \pi+\omega^*], \alpha \in [0, \tan^{-1}(\sqrt{s}r\cos \theta) + \frac{\pi}{2}], \\
        \text{For } \omega &\in [\pi+\omega^*, 2\pi], \alpha \in [0, \tan^{-1}(\sqrt{n_1-s}r\sin \theta) + \frac{\pi}{2}],
    \end{align*}
    where $\tan \omega^* = \frac{s}{n_1-s}$.
    By uniformly integrating the above domain in $\mathbb{S}^2$,
    we have 
    \begin{align*}
        \mathfrak{p}_2(s) = \frac{1}{2} + \left[\int_{\frac{\pi}{2}}^{\pi+\alpha_{s}} \frac{g(r\sqrt{s}\cos(\theta))}{4\pi}d\theta + \int_{\pi+\alpha_{s}}^{2\pi} \frac{g(r\sqrt{n_1-s}\sin(\theta))}{4\pi}d\theta \right],
    \end{align*}
    where $g(x) = \sin(\tan^{-1}(x))$.
    Thus, we obtain
    $$            (\text{P}_2)_{n_1+1,j+1}=\binom{n_2}{j}\mathbb{E}_{s}\left[(1-\mathfrak{p}_2(s))^j\mathfrak{p}_2(s)^{n_2-j}\right], \quad 0 \le j \le n_2.
    $$
    
    Secondly, let us consider the case where  
    the `unit hypersphere' initialization with bias is used for the 1st hidden layer and $n_1=1$. 
    Since $[w_1,b_1] \sim \mathbb{S}^1$, 
    we write it as $(\sin\omega,\cos\omega)$
    for $\omega \in [-\pi,\pi]$.
    Since $x \in [-r,r]$, 
    we have 
    \begin{align*}
        A_1^0 &= \{\omega \in [-\pi,\pi] | \phi(\sin\omega x + \cos \omega)=0, \forall x \in [-r,r] \} = [-\pi+\alpha_r,\pi-\alpha_r], \\
        A_1^1 &= (A_1^0)^c = (-\pi+\alpha_r, \pi-\alpha_r),
    \end{align*}
    where $\alpha_r = \tan^{-1}(r)$.
    If the `normal' initialization without bias is used for the 2nd hidden layer,
    since a single neuron in the 2nd layer is
    $\phi(w^2\phi(w^1x+b^1))$, 
    for given $A_1^1$, we have $\mathfrak{p}_2(A_1^1) = \frac{1}{2}$.
    Thus,
    $$
    \text{Pr}(\text{m}_2=j|\text{m}_1=1) =
    \binom{n_2}{j}(1/2)^j(1/2)^{n_2-j}, \quad j=0,\cdots,n_2.
    $$
    If the `normal' initialization with bias is used for the 2nd hidden layer,
    it follows from Lemma~\ref{lem:bdp-2nd}
    that for $\omega \in A_1^1$,
    \begin{align*}
    \mathfrak{p}_2(\omega)
    =
    \begin{cases}
    \frac{1}{4} + \frac{g(|\omega| - \alpha_r)}{2\pi}, 
    &\text{if } 
    |\omega| \in \left[\frac{\pi}{2}-\alpha_r, \frac{\pi}{2}+\alpha_r\right), \\
    \frac{1}{4} +\frac{g(|\omega| - \alpha_r) + \tan^{-1}\left(\sqrt{r^2+1}\cos(|\omega| + \alpha_r)\right)}{2\pi},
    & \text{if }
    |\omega| \in \left[0, \frac{\pi}{2}-\alpha_r\right),
    \end{cases}
    \end{align*}
    where $g(x) = \tan^{-1}\left(\frac{1}{\sqrt{r^2+1}\cos(x)}\right)$.
    
    Thus, we have
    $$
    \text{Pr}(\text{m}_2=j|\text{m}_1=1) =
    \binom{n_2}{j}\mathbb{E}_\omega\left[(1-\mathfrak{p}_2(\omega))^j(\mathfrak{p}_2(\omega))^{n_2-j}\right], \quad j=0,\cdots,n_2,
    $$
    where $\omega \sim \text{Unif}(A_1^1)$.
    Then the proof is completed once we have the following lemma.
    
    \begin{lemma} \label{lem:bdp-2nd}
        Given a ReLU network having $\bm{n}=(1,1,n_2,\cdots,n_L)$,
        suppose $(w^1,b^1), (w^2,b^2)\sim \mathbb{S}^1$.
        Given $\{w^1,b^1\}$, let $\omega$ be the angle of $(w^1,b^1)$ in $\mathbb{R}^2$.
        Then, the BDP for a ReLU neuron at the 2nd hidden layer is 
        \begin{align*}
        \mathfrak{p}_2(\omega)
        =
        \begin{cases}
        1, 
        & \text{if } 
        |\omega| \in \left[\frac{\pi}{2}+\alpha_r, \pi\right], \\
        \frac{1}{4} + \frac{g(|\omega| - \alpha_r)}{2\pi}, 
        &\text{if } 
        |\omega| \in \left[\frac{\pi}{2}-\alpha_r, \frac{\pi}{2}+\alpha_r\right), \\
        \frac{1}{4} +\frac{g(|\omega| - \alpha_r) + \tan^{-1}\left(\sqrt{r^2+1}\cos(|\omega| + \alpha_r)\right)}{2\pi},
        & \text{if }
        |\omega| \in \left[0, \frac{\pi}{2}-\alpha_r\right), 
        \end{cases}
        \end{align*}
        where $g(x) = \tan^{-1}\left(\frac{1}{\sqrt{r^2+1}\cos(x)}\right)$
        and $\alpha_r = \tan^{-1}(r)$.
    \end{lemma}
    \begin{proof}[Proof of Lemma~\ref{lem:bdp-2nd}]
        For a fixed $\bm{v}=[w,b]$ and $\bm{z}=[x,1]$, we can write 
    \begin{align*}
    \phi(\bm{v}^T\bm{z}) = \|\bm{z}\|\phi(\bm{v}^T\bm{z}/\|\bm{z}\|) =\|z\|\phi(\cos(\omega - \theta(x))), \quad \theta(x) = \tan^{-1}(x).
    \end{align*}
    Since $\bm{v}$ is uniformly drawn from $\mathbb{S}^1$, it is equivalent to draw 
    $\omega \sim \mathcal{U}(-\pi,\pi)$.
    Let $0 < \theta_{\max} = \tan^{-1}(r) < \pi/2$.
    Then,
    \begin{align*}
    \phi(\bm{v}^T\bm{z}) =
    \begin{cases}
    \bm{v}^T\bm{z},  & \forall \theta(x), \text{ if } \omega \in \left(-\frac{\pi}{2}+\theta_{\max}, \frac{\pi}{2}-\theta_{\max}\right), \\
    0,  & \forall \theta(x), \text{ if } \omega \in \left[-\pi, \-\frac{\pi}{2}-\theta_{\max}\right]\cup\left[\frac{\pi}{2}+\theta_{\max}, \pi\right],
    \end{cases}
    \end{align*}
    and if $\omega \in \left(-\frac{\pi}{2} -\theta_{\max}, -\frac{\pi}{2} +\theta_{\max} \right] \cup \left[\frac{\pi}{2} -\theta_{\max}, \frac{\pi}{2} +\theta_{\max} \right)$,
    we have
    $\phi(\bm{v}^T\bm{z}) = \bm{v}^T\bm{z}\mathbb{I}_{A(\omega)}(\theta(x))$,
    where
    $A(\omega) = \{\theta \in [-\theta_{\max},\theta_{\max}] | |\theta(x)-\omega| \le \frac{\pi}{2} \}$.
    Due to symmetry, let us assume that $\omega \sim \mathcal{U}(0,\pi)$.
    Then, it can be checked that $
    {A}_1^0=\left[\frac{\pi}{2}+\theta_{\max}, \pi\right]$
    and ${A}_1^1=\left[0,\frac{\pi}{2}+\theta_{\max}\right)$.
    Furthermore,
    \begin{align*}
    \max_{\bm{z}} \phi(\bm{v}^T\bm{z}) =
    \begin{cases}
    \sqrt{r^2+1}\cos(\omega - \theta_{\max}) , & \text{if } \omega \in \left(0, \frac{\pi}{2} +\theta_{\max} \right),  \\
    0, & \text{if } \omega \in \left[\frac{\pi}{2}+\theta_{\max}, \pi\right], 
    \end{cases}
    \end{align*}
    and
    \begin{align*}
    \min_{\bm{z}} \phi(\bm{v}^T\bm{z}) =
    \begin{cases}
    \sqrt{r^2+1}\cos(\omega + \theta_{\max}) , & \text{if } \omega \in \left(0, \frac{\pi}{2}-\theta_{\max}\right), \\
    0, & \text{if } \omega \in \left[\frac{\pi}{2}-\theta_{\max}, \pi\right]. 
    \end{cases}
    \end{align*}
    For a fixed $\omega$,
    let $\mathfrak{p}_2(\omega)$ be the probability that a single neuron at the 2nd layer is born dead, i.e., 
    $$
    \mathfrak{p}_2(\omega) = \text{Pr}(w^2\phi(w^1x+b^1) + b^2 < 0, \forall x \in B_r(0) | w^1, b^1).
    $$
    Also, since $(w^2,b^2) \overset{d}{=} (-w^2,-b^2)$, we have
    $$
    \mathfrak{p}_2(\omega) = \text{Pr}(w^2\phi(w^1x+b^1) + b^2 > 0, \forall x \in B_r(0) | w^1, b^1).
    $$
    It follows from \eqref{eqn-pd1} that
    $$
    \mathfrak{p}_2(\omega) = \frac{1}{4} + \frac{\tan^{-1}(1/\max_{\bm{z}} \phi(\bm{v}^T\bm{z})) + \tan^{-1}(\min_{\bm{z}} \phi(\bm{v}^T\bm{z}))}{2\pi}.
    $$
    Thus, we obtain
    \begin{align*}
    \mathfrak{p}_2(\omega)
    =
    \begin{cases}
    1, 
    & \text{if } 
    \omega \in \left[\frac{\pi}{2}+\theta_{\max}, \pi\right], \\
    \frac{1}{4} + \frac{g(\omega - \theta_{\max})}{2\pi}, 
    &\text{if } 
    \omega \in \left[\frac{\pi}{2}-\theta_{\max}, \frac{\pi}{2}+\theta_{\max}\right), \\
    \frac{1}{4} +\frac{g(\omega - \theta_{\max}) + \tan^{-1}\left(\sqrt{r^2+1}\cos(\omega + \theta_{\max})\right)}{2\pi},
    & \text{if }
    \omega \in \left[0, \frac{\pi}{2}-\theta_{\max}\right), 
    \end{cases}
    \end{align*}
    where $g(x) = \tan^{-1}\left(\frac{1}{\sqrt{r^2+1}\cos(x)}\right)$
    and this completes the proof.
    \end{proof}
\end{proof}

\begin{figure}[!htbp]
	\centerline{
		\includegraphics[width=5.0cm]{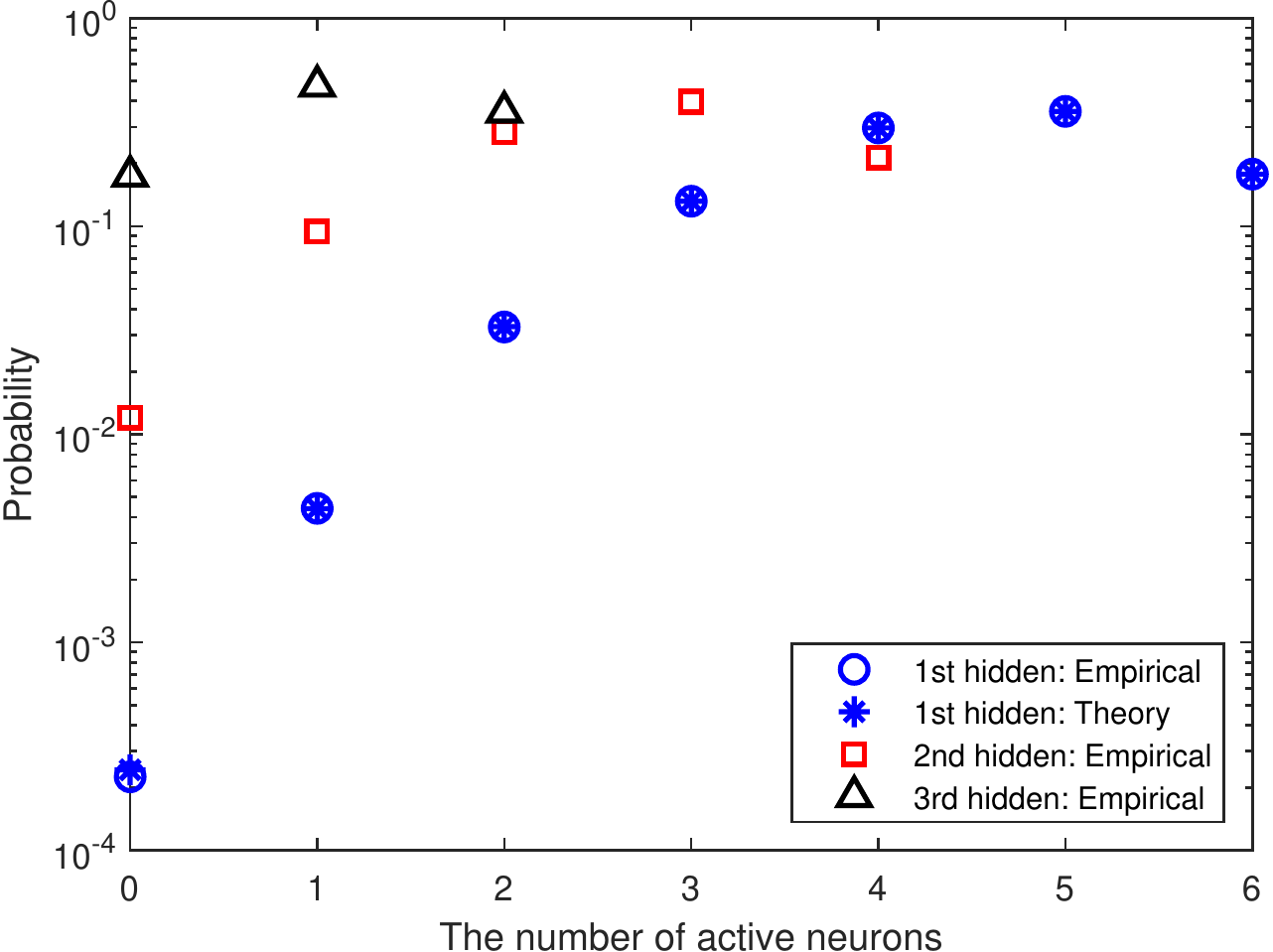}
		\includegraphics[width=5.0cm]{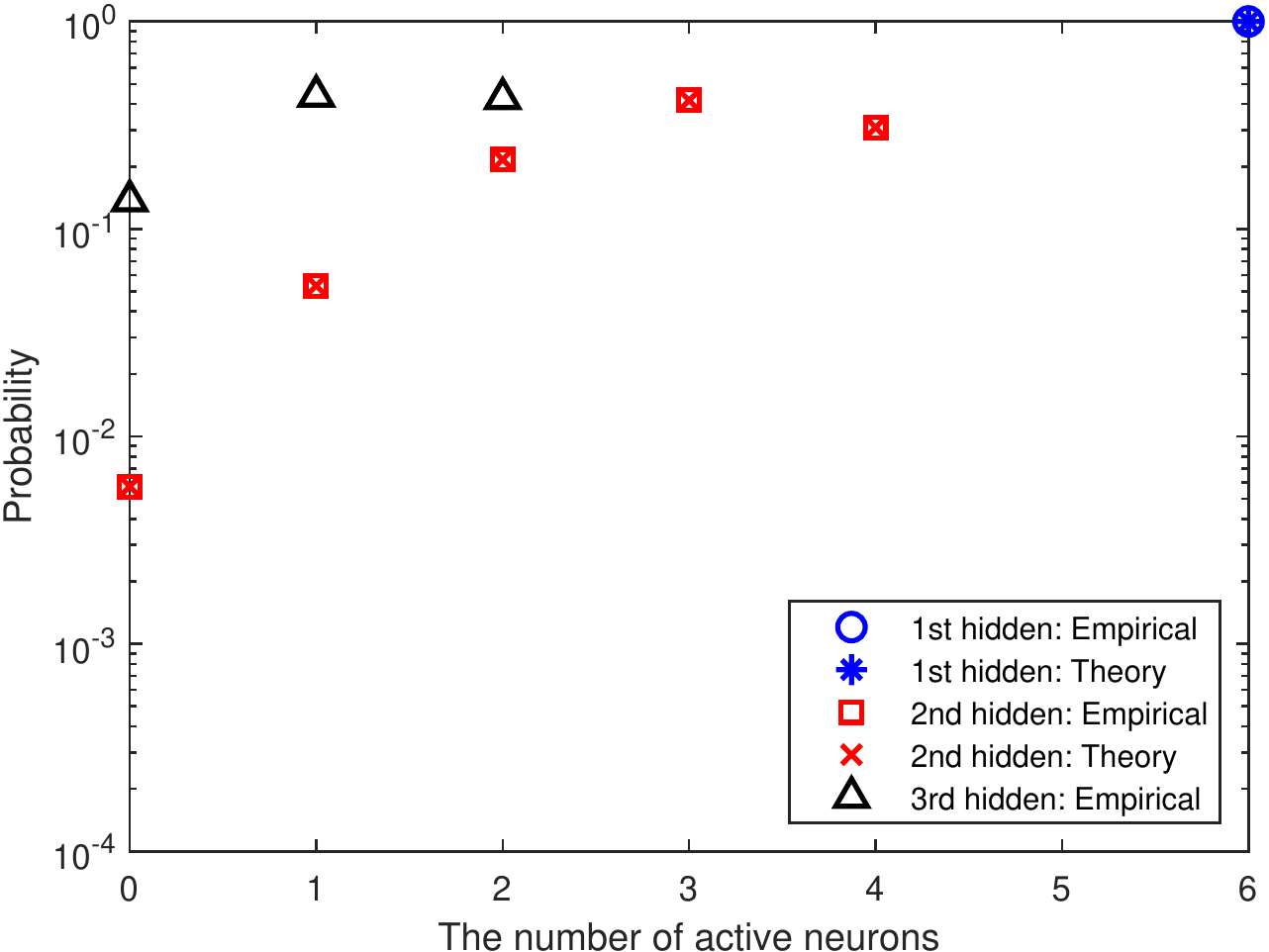}
		\includegraphics[width=5.0cm]{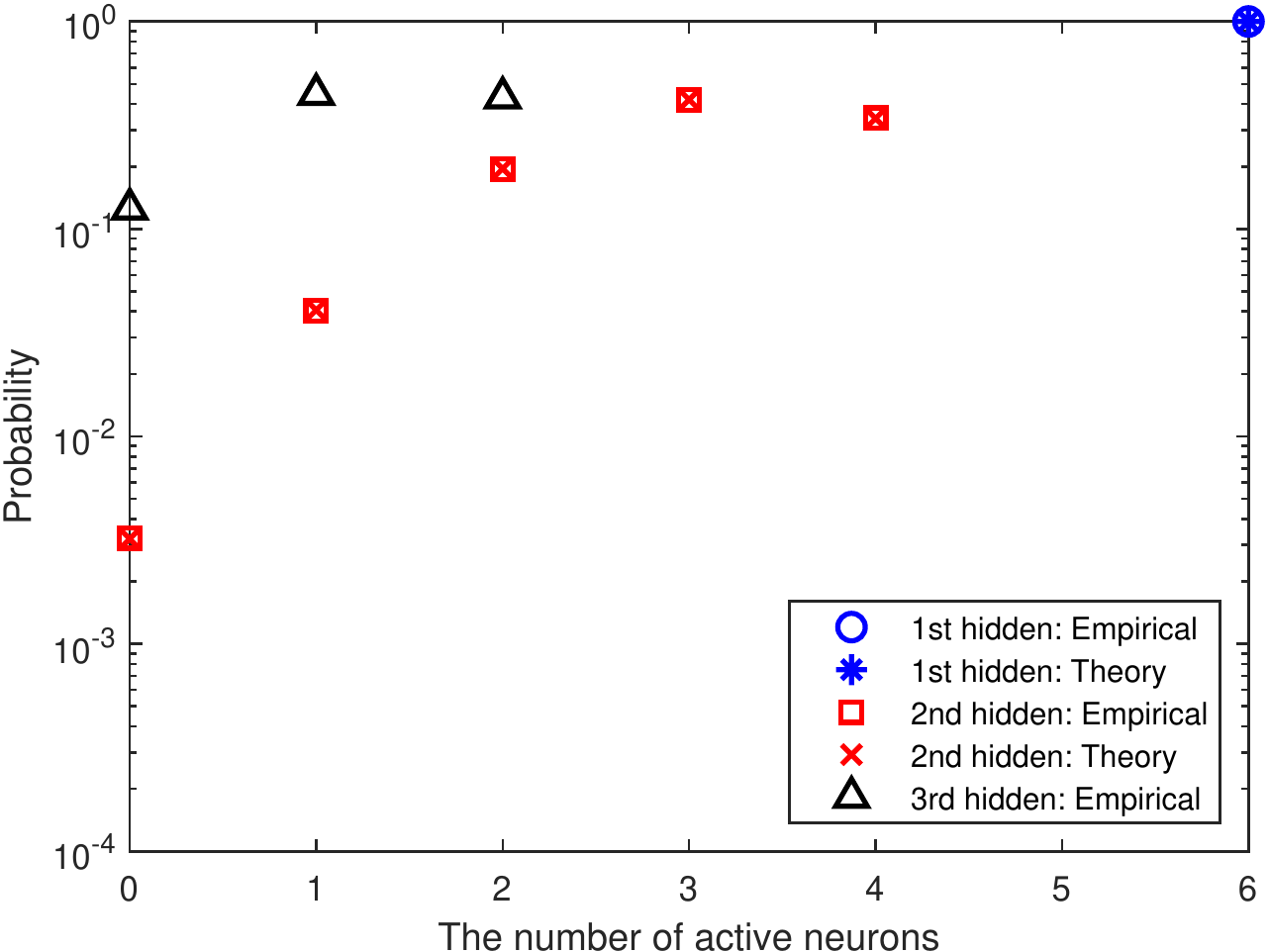}
	}
	\caption{The probability distributions of the number of active neurons at different layers are shown for a ReLU network having $\bm{n}=(1,6,4,2,n_4,\cdots,n_L)$ architecture. 
		(Left) All layers are initialized by the `unit hypersphere' with bias. 
		(Middle) All layers are initialized by the `unit hypersphere' without bias.
		(Right) The first hidden layer is initialized by the `unit hypersphere' without bias. All other layers are initialized by the `normal' with bias.}
	\label{fig:pi_j-1642}
\end{figure}

Lemmas~\ref{THM:PI_1} and~\ref{THM:PI_2} indicate that 
the bias initialization could drastically change the active neuron distributions $\pi_j$.
Since $\pi_j = \pi_1\text{P}_2\cdots \text{P}_j = \pi_2\text{P}_3\cdots \text{P}_j$, 
the behaviors of $\pi_1$ and $\pi_2$ 
affect the higher layer's distributions $\pi_j$.
In Figure~\ref{fig:pi_j-1642}, we consider a ReLU network with $\bm{n}=(1,6,4,2,n_4,\cdots,n_L)$ architecture
and plot the empirical distributions $\pi_j$, $j=1,2,3$, from $10^6$ independent simulations at $r=1$.
On the left and the middle, the `unit hypersphere' \eqref{initialization-nsphere} initialization without and with bias are employed, respectively, in all layers.
%
On the right, the `unit hypersphere' initialization without bias is employed in the 1st hidden layer,
and the `normal' \eqref{initialization-He} initialization with bias is employed in all other layers. 
The theoretically derived distributions, $\pi_1, \pi_2$, are also plotted as references.
We see that 
all empirical results are well matched with our theoretical derivations. 
When the 1st hidden layer is initialized with bias, with probability 0.8, at least one neuron in the 1st hidden layer will be dead. 
On the other hand, if the 1st hidden layer is initialized without bias, with probability 1, no neuron will be dead. 
It is clear that the distributions obtained by three initialization schemes 
show different behavior.

\section{A general formulation for computing trainability} \label{sec:gen-formula-trainability}
We present a general formulation for computing trainability.
Our formulation requires a complete understanding of two types of inhomogeneous stochastic matrices. 

Let $\mathfrak{d}_t^b$ be the number of permanently dead neurons at the $t$-th hidden layer.
Given $\{n_t,m_t\}_{t=1}^{L-1}$, 
let $s_t = (n_t-m_t+1)(m_t-1)$ for $t > 1$ and $s_1 = n_1-m_1+1$.
For convenience, let $\mathcal{T}_{t-1} := [\hat{n}_1]\times [\hat{n}_2]\times [\hat{m}_2]\cdots \times [\hat{n}_{t-1}]\times [\hat{m}_{t-1}]$,
where $[\hat{n}_t] = \{0,\cdots, n_t - m_t\}$
and $[\hat{m}_t] = \{1,\cdots,m_t-1\}$.
Let $\hat{T}_{t} := [\hat{n}_{t}]\times [\hat{m}_{t}]$.
Let 
$$
\text{k}_{t-1}^l = (k^l_1,k^l_2, k^l_{2,b},\cdots,k_{t-1}^l,k_{t-1,b}^l)
$$ 
be the $l$-th multi-index of 
$\mathcal{T}_{t-1}$ (assuming a certain ordering).
For $1 < t < L-1$, 
let $\hat{P}_t$ be a matrix of size $\prod_{j=1}^{t-1} s_j \times \prod_{j=1}^{t} s_j$ defined as follow.
For $l=1,\dots,\prod_{j=1}^{t-1} s_j$ and $r = 1,\dots, \prod_{j=1}^{t} s_j$,
\begin{equation} \label{def:stochasticMAT-type2}
[\hat{P}_t]_{l,r} = \text{Pr}(\m_t = k^r_t, \mathfrak{d}_t^b = k^r_{t,b} |\m_s = k^l_s, \mathfrak{d}_s^b = k^l_{s,b}, \forall 1\le s < t) 
\prod_{j=1}^{t-1}\delta_{k^l_j = k^r_j}\delta_{k^l_{j,b} = k^r_{j,b}}.
\end{equation}
For $t=L-1$, 
let $\hat{P}_{L-1}$ be a matrix of size $\prod_{j=1}^{L-2} s_j \times s_{L-1}$ such that
for $l=1,\dots,\prod_{j=1}^{L-2} s_j$ and $r=1,\dots,s_{L-1}$, 
\begin{equation} \label{def:stochasticMAT-type2-L}
[\hat{P}_{L-1}]_{l,r} = \text{Pr}(\m_{L-1} = \bar{k}_{L-1}^r, \mathfrak{d}_{L-1}^b = \bar{k}_{L-1,b}^r |\m_s = k^l_s, \mathfrak{d}_s^b = k^l_{s,b}, \forall 1\le s < L-2),
\end{equation}
where $\bar{\text{k}}_{L-1}^r = (\bar{k}_{L-1}^r, \bar{k}_{L-1,b}^r)$ is the $r$-th multi-index of the lexicographic ordering of $[\hat{n}_{L-1}]\times [\hat{m}_{L-1}]$.

Once the above stochastic matrices are all identified, 
its corresponding trainability readily follows 
based on the formulation given below.
\begin{lemma} \label{lem:gen-formula-trainability}
	For a learning task that requires a $L$-layer ReLU network having at least $m_t$ active neurons in the $t$-th layer,
	the trainability for a $L$-layer ReLU network with $\bm{n}=(n_0,n_1,\cdots,n_L)$ architecture is given as follow.
	Let $\tilde{n}_t = n_t - m_t+1$.
	Then, the trainability is given by
	\begin{align*}
	\text{Trainability } = \pi'_1 \text{P}'_2 \cdots \text{P}'_{L-1}\mathbb{1}_{\tilde{n}_{L-1}} + {\pi}'_1 \hat{P}_2 \cdots \hat{P}_{L-1} \mathbb{1}_{s_{L-1}},
	\end{align*}
	where $\pi'_1$ is a $1\times \tilde{n}_1$ submatrix of $\pi_1$ whose first component is $[\pi_1]_{m_t}$,
	$\text{P}'_t$ is a $\tilde{n}_{t-1} \times \tilde{n}_t$ submatrix of $\text{P}_t$ whose $(1,1)$-component is $[\text{P}_t]_{m_{t-1},m_t}$,
	and $\mathbb{1}_p$ is a $p \times 1$ vector whose entries are all 1s.
	Here $\pi_1$ and $\text{P}_t$ are defined in Lemma~\ref{def:ANPDist},
	and $\{\hat{P}_t\}$ is defined in \eqref{def:stochasticMAT-type2} and \eqref{def:stochasticMAT-type2-L}. 
\end{lemma}
\begin{proof}[Proof of Lemma~\ref{lem:gen-formula-trainability}]
	We observe that 
	\begin{align*}
	&\text{Pr}(\m_t \ge 1,  \mathfrak{d}_{t}^b \le n_t - m_t, \forall 1\le t < L) \\
	&=\text{Pr}(\m_t \ge m_t, \forall 1 \le t < L) 
	+ \text{Pr}(\m_1 \ge m_1, 1\le \m_t < m_t, \mathfrak{d}_{t}^b \le n_t - m_t, \forall 1\le t < L).
	\end{align*}
	From Lemma~\ref{def:ANPDist}, it can be checked that
	$$
	\text{Pr}(\m_t \ge m_t, \forall 1 \le t < L) = \pi'_1 \text{P}'_2 \cdots \text{P}'_{L-1}\mathbb{1}_{\tilde{n}_{L-1}}.
	$$ 
	For convenience, let 
	$\hat{\m}_t = (\m_t, \mathfrak{d}_t^b)$ for $t > 1$ 
	and $\hat{\m}_1 = \m_t$.
	Let $\vec{\m}_t = (\hat{\m}_1,\hat{\m}_2,\cdots,\hat{\m}_t)$.
	Also, recall that $\mathcal{T}_{t-1} := [\hat{n}_1]\times [\hat{n}_2]\times [\hat{m}_2]\cdots \times [\hat{n}_{t-1}]\times [\hat{m}_{t-1}]$,
	where $[\hat{n}_t] = \{0,\cdots, n_t - m_t\}$
	and $[\hat{m}_t] = \{1,\cdots,m_t-1\}$.
	Let $\hat{T}_{t} := [\hat{n}_{t}]\times [\hat{m}_{t}]$.
	Also let $\vec{\pi}_t = [\text{Pr}(\vec{\m}_{t} = \text{k})]_{\text{k} \in \mathcal{T}_t}$ be the distribution of $\vec{\m}_t$ restricted to $\mathcal{T}_t$.
	Then, 
	\begin{align*}
	&\text{Pr}(\m_1 \ge m_1, 1\le \m_t < m_t, \mathfrak{d}_{t}^b \le n_t - m_t, \forall 1\le t < L) \\
	&=\text{Pr}(\vec{\m}_{L-1} \in \mathcal{T}_{L-1})
	= \sum_{\text{k}_{L-1} \in \mathcal{T}_{L-1}}  \text{Pr}(\vec{\m}_{L-1} = \text{k}_{L-1})
	\\
	&= 
	\sum_{\hat{\text{k}}_{L-1} \in \hat{T}_{L-1}} 
	\sum_{\text{k}_{L-2} \in \mathcal{T}_{L-2}} \text{Pr}(\hat{\m}_{L-1} =\hat{\text{k}}_{L-1} | \vec{\m}_{L-2} = \text{k}_{L-2}) \text{Pr}(\vec{\m}_{L-2} = \text{k}_{L-2})
	\\
	&= \vec{\pi}_{L-2} \hat{P}_{L-1}\mathbb{1}_{s_{L-1}}.
	\end{align*}
	It then suffices to identify $\vec{\pi}_t$ for $1\le t < L-1$.
	Then, note that
	for each $\text{k}_t = (\text{k}_{t-1},\hat{\text{k}}_t) \in \mathcal{T}_t$,
	\begin{align*}
		\text{Pr}(\vec{\m}_{t} = \text{k}_t)
		= \text{Pr}(\hat{\m}_t = \hat{\text{k}}_t |\vec{\m}_{t-1} = \text{k}_{t-1})\text{Pr}(\vec{\m}_{t-1} = \text{k}_{t-1})
	\end{align*}
	Thus, we have
	$\vec{\pi}_t = \vec{\pi}_{t-1}\hat{P}_{t}$.
	Since $\vec{\pi}_1 = \pi'_1$, by recursively applying it,
	the proof is completed.
\end{proof}

We are now in a position to present our proof of Theorem~\ref{THM:DEEP}.
\begin{proof}[Proof of Theorem~\ref{THM:DEEP}]
	Given the event $A_{t-1}^i$ that exactly $i$ neurons are active in the $(t-1)$-th hidden layer, 
	let 
	$\mathfrak{p}_{t,b}(A_{t-1}^i)$
	and $\mathfrak{p}_{t,g}(A_{t-1}^i)$
	be the conditional probabilities that a neuron in the $t$-th hidden layer is born dead permanently
	and born dead tentatively, respectively.
	Then, 
	$$
	\mathfrak{p}_t(A_{t-1}^i) = \mathfrak{p}_{t,b}(A_{t-1}^i)
	+ \mathfrak{p}_{t,g}(A_{t-1}^i).
	$$
	Note that since the weights and the biases are initialized from symmetric probability distribution around 0,
	we have $\mathfrak{p}_{t,b}(A_{t-1}^i) \ge 2^{-i-1}$.
	This happens when all the weights and bias are initialized to be non-positive.
	Let $\mathfrak{d}_t^g$ and $\mathfrak{d}_t^b$ be the number of tentatively dead and permanently dead neurons 
	at the $t$-th hidden layer.
	It then can be checked that 
	\begin{align*}
	&\text{Pr}(\mathfrak{d}_t^g=j_1, \mathfrak{d}_t^b=j_2 | \m_1=i) \\
	&= \binom{n_t}{j_1,j_2,j_3} 
	\mathbb{E}_{t-1}
	\left[
	(1-\mathfrak{p}_t(A_{t-1}^i))^{n_t-j_1-j_2}
	(\mathfrak{p}_{t,g}(A_{t-1}^i))^{j_1}
	(\mathfrak{p}_{t,b}(A_{t-1}^i))^{j_2}
	\right],
	\end{align*}
	where $j_3=n_t-j_1-j_2$, $\mathbb{E}_{t-1}$ is the expectation with respect to $\mathcal{F}_{t-1}$
	and $\binom{n}{k_1,k_2,k_3}$ is a multinomial coefficient.
	Also note that $\m_t +\mathfrak{d}_t^g+\mathfrak{d}_t^b = n_t$.
	It then follows from Lemma~\ref{lem:gen-formula-trainability} that
	\begin{align*}
	&\text{Pr}(\m_t \ge 1,  \mathfrak{d}_{t}^b \le n_t - m_t, \forall 1\le t < 3) \\
	&= \text{Pr}(\m_1 \ge m_1, \m_2 \ge m_2)
	+ \sum_{j=1}^{m_2-1}\sum_{l=0}^{n_2-m_2}\sum_{k=m_1}^{n_1} \text{Pr}(\m_2 = j,  \mathfrak{d}_{2}^b = l |\m_1 = k)\text{Pr}(\m_1 = k) \\
	&= \text{Pr}(\m_1 \ge m_1, \m_2 \ge m_2)
	+ \sum_{j=1}^{m_2-1}\sum_{l=0}^{n_2-m_2}\sum_{k=m_1}^{n_1} \text{Pr}(\mathfrak{d}_2^g=n_2- j-l,  \mathfrak{d}_{2}^b = l |\m_1 = k)\text{Pr}(\m_1 = k) \\
	&= \text{Pr}(\m_1 \ge m_1, \m_2 \ge m_2) \\
	&\qquad+ \sum_{j=1}^{m_2-1}\sum_{l=0}^{n_2-m_2}\sum_{k=m_1}^{n_1} 
	\binom{n_2}{n_2-j-l,j,l} \mathbb{E}_{1}\left[(1-\mathfrak{p}_2(A_{1}^k))^{j}
	(\mathfrak{p}_{2,g}(A_{1}^k))^{n_2-j-l}
	(\mathfrak{p}_{2,b}(A_{1}^k))^{l}
	\right]\text{Pr}(\m_1 = k) \\
	&\ge \text{Pr}(\m_1 \ge m_1, \m_2 \ge m_2) \\
	&\qquad+ \sum_{j=1}^{m_2-1}\sum_{l=0}^{n_2-m_2}\sum_{k=m_1}^{n_1} 
	\binom{n_2}{n_2-j-l,j,l} \mathbb{E}_{1}\left[(1-\mathfrak{p}_2(A_{1}^k))^{j}
	(\mathfrak{p}_2(A_{1}^k) - 2^{-k-1})^{n_2-j-l}
	(2^{-k-1})^{l}
	\right]\text{Pr}(\m_1 = k).
	\end{align*}
	Since $\mathfrak{p}_2(A_{1}^k)$ is identified by Lemma~\ref{THM:PI_2}
	and $\text{Pr}(\m_1 = k)$ is identified by Lemma~\ref{THM:PI_1}, 
	by plugging it on the above, the proof is completed.
\end{proof}

\section{Proof of Corollary~\ref{COR:TRAINABILITY-UPPER-1D}} \label{app:COR:TRAINABILITY-UPPER-1D}
\begin{proof}
    Note that 
    $$
    \text{Pr}(\m_t \ge 1,  \mathfrak{d}_{t}^b \le n_t - m_t, \forall t=1,\cdots,L)
    \le 
    \text{Pr}(\m_t \ge 1, \forall t=1,\cdots,L),
    $$
    and
    $$
    1 - \text{Pr}(\m_t \ge 1, \forall t=1,\cdots,L)
    = \text{Pr}(\exists t, \text{ such that } \m_t = 0)
    = \text{Pr}(\N^{L+1}(\x) \text{ is born dead}).
    $$
    It was shown in Theorem 3 of \cite{lu2019dying} that 
    $$
    \text{Pr}(\N^L(\x) \text{ is born dead})
    \ge 1 - \mathfrak{a}_1^{L-2} +
        \frac{(1-2^{-n+1})(1-2^{-n})}{1+(n-1)2^{-n}}(-\mathfrak{a}_1^{L-2} + \mathfrak{a}_2^{L-2}),
    $$
    where
    $\mathfrak{a}_1 = 1-2^{-n}$
    and $\mathfrak{a}_2 = 1-2^{-n+1}-(n-1)2^{-2n}$.
    Thus, the proof is completed.
\end{proof}

\section{Proof of Theorem~\ref{THM:DATA-DEPENDENT}} \label{app:THM:DATA-DEPENDENT}
\begin{proof} 
Since $q(\x) = E[\|\N(\x)\|^2]/d_\text{out}$ and the rows of $\bm{W}^2$ are independent, without loss of generality, 
let us assume $d_\text{out}= 1$. 
The direct calculation shows that 
\begin{align*}
E[\|\N(\x_k)\|^2] &= \sum_{i=1}^N \sigma_\text{out}^2 E\left[ \phi(\bm{w}^T_i\x_k + \bm{b}_i)^2 \right] \\
&= 
\frac{N\sigma_\text{out}^2}{N_\text{train}}\left(
\sum_{i=1}^{N_\text{train}} E\left[ \phi(\bm{w}^T_i(\x_k-\x_{i}) + |\epsilon_i|)^2 \right]\right),
\end{align*}
Let $\sigma_{k,i}^2 = \sigma_\text{in}^2 \|\x_k - \x_i\|^2$
and $\epsilon_{k,i} = |\epsilon_i|/\sigma_{k,i}$.
Note that $\bm{w}_i^T(\x_k-x_i) \sim N(0,\sigma_{k,i}^2)$.
Then,
\begin{align*}
    E\left[ \phi(\bm{w}^T_i(\x_k-\x_{i}) + |\epsilon_i|)^2 | \epsilon_i \right]  = I_1(\epsilon_i) + I_2(\epsilon_i),
\end{align*}
where
$$
I_1(\epsilon) = \int_{0}^\infty (z+\epsilon)^2 \frac{e^{-\frac{z^2}{2\sigma_{k,i}^2}}}{\sqrt{2\pi \sigma_{k,i}^2}} dz, \qquad 
I_2(\epsilon) = \int_{-\epsilon}^0 (z+\epsilon)^2 \frac{e^{-\frac{z^2}{2\sigma_{k,i}^2}}}{\sqrt{2\pi \sigma_{k,i}^2}} dz.
$$
Then, if 
$\epsilon_{i} = |e_{i}|$ where $e_{i} \sim \text{N}(0,\sigma_{e,i}^2)$,
we have
$$
I_1(\epsilon_{i}) = \frac{1}{2}\sigma_{k,i}^2 + \sqrt{\frac{2}{\pi}}
\sigma_{k,i} \epsilon_i + \frac{1}{2}\epsilon_i^2
\implies 
E[I_1(\epsilon_i)] = \frac{1}{2}\sigma_{k,i}^2 + \frac{2}{\pi}\sigma_{k,i}\sigma_{e,i} + \frac{1}{2}\sigma_{e,i}^2.
$$
Also, we have
\begin{align*}
I_2(\epsilon)
&=\int_{-\epsilon}^0 (z+\epsilon)^2 \frac{e^{-\frac{z^2}{2\sigma_{k,i}^2}}}{\sqrt{2\pi \sigma_{k,i}^2}} dz \\
&= \sigma_{k,i}^2 \int_{-\epsilon_{k,i}}^0 (z+\epsilon_{k,i})^2 \frac{e^{-\frac{z^2}{2}}}{\sqrt{2\pi }} dz \\
&=\sigma_{k,i}^2\left(
\frac{1}{2}(\epsilon_{k,i}^2 + 1) \text{erf}\left(\frac{\epsilon_{k,i}}{\sqrt{2}}\right)
+ \frac{\epsilon_{k,i}(e^{-\epsilon_{k,i}^2/2}-2)}{\sqrt{2\pi}}
\right),
\end{align*}
where
$$
\epsilon_{k,i} = |e_{k,i}|, \qquad \text{where} \qquad e_{k,i} \sim \text{N}(0,\sigma_{e,i}^2/\sigma_{k,i}^2).
$$
Note that if $z = |z'|$ where $z' \sim \text{N}(0,\sigma^2)$,
\begin{align*}
E[z^2\text{erf}(z/\sqrt{2})] &= \frac{2\sigma^2\tan^{-1}(\sigma)}{\pi} + \frac{2\sigma^3}{\pi(\sigma^2+1)}, \\
E[\text{erf}(z/\sqrt{2})] &= \frac{2\tan^{-1}(\sigma)}{\pi}, \\
E[z e^{-z^2/2}] &= \frac{2\sigma}{\sqrt{2\pi}(\sigma^2+1)}, \\
E[z] &= \frac{2\sigma}{\sqrt{2\pi}}.
\end{align*}
Therefore,
\begin{align*}
E\left[\frac{1}{2}(z^2+1)\text{erf}(z/\sqrt{2}) + \frac{ze^{-z^2/2}-2z}{\sqrt{2\pi}} \right]
&=
\frac{(\sigma^2+1)\tan^{-1}(\sigma)}{\pi}-\frac{\sigma}{\pi}.
\end{align*}
By setting $\sigma = \sigma_{e,i}/\sigma_{k,i}$, we have
\begin{align*}
    E[I_2(\epsilon_i)] &= \sigma_{k,i}^2\mathbb{E}_{\epsilon_{k,i}}\left[
\frac{1}{2}(\epsilon_{k,i}^2 + 1) \text{erf}\left(\frac{\epsilon_{k,i}}{\sqrt{2}}\right)
+ \frac{\epsilon_{k,i}(e^{-\epsilon_{k,i}^2/2}-2)}{\sqrt{2\pi}}
\right],
    \\
    &= \frac{(\sigma_{e,i}^2+\sigma_{k,i}^2)\tan^{-1}(\sigma_{e,i}/\sigma_{k,i})}{\pi} - \frac{\sigma_{e,i}\sigma_{k,i}}{\pi}:=\gamma_i.
\end{align*}
Thus, we have
\begin{align*}
    E\left[ \phi(\bm{w}^T_i(\x_k-\x_{i}) + |\epsilon_i|)^2 \right]  &= E[I_1(\epsilon_i)] + E[I_2(\epsilon_i)]
    = \frac{1}{2}\sigma_{k,i}^2 + \frac{2}{\pi}\sigma_{k,i}\sigma_{e,i} + \frac{1}{2}\sigma_{e,i}^2 +\gamma_i,
\end{align*}
and thus,
\begin{equation*}
   E[\|\N(\x_k)\|^2]  =\frac{N\sigma_\text{out}^2}{N_\text{train}}
\sum_{i =1}^{N_\text{train}}
\left[\frac{1}{2}\sigma_{k,i}^2 + \frac{1}{2}\sigma_{e,i}^2 + \frac{2}{\pi}\sigma_{k,i}\sigma_{e,i} + \gamma_i \right] .
\end{equation*}
Let $\sigma_{e,i}^2 = \sigma_e^2= \sigma_\text{in}^2s^2$ for all $i$. Then we have
\begin{equation*}
    E[q(\x_k)] = E[\|\N(\x_k)\|^2] 
    =
\frac{N\sigma^2_\text{out}\sigma_\text{in}^2}{N_\text{train}\pi}\sum_{i=1}^{N_\text{train}}
\left[(s^2+\Delta_{k,i}^2)\left(\tan^{-1}(s/\Delta_{k,i})+\pi/2\right)+s\Delta_{k,i} \right],
\end{equation*}
where $\Delta_{k,i} = \|\x_k-\x_i\|_2$.
Thus,
we obtain 
\begin{equation*}
    \mathbb{E}_{\mathcal{X}_m}[q(\x)] = \frac{1}{N_\text{Ntrain}}\sum_{k=1}^{N_\text{train}}
E[q(\x_k)] 
=\frac{N\sigma^2_\text{out}\sigma_{\text{in}}^2}{N_\text{train}^2\pi}\sum_{k,i=1}^{N_\text{train}}
\left[(s^2+\Delta_{k,i}^2)\left(\tan^{-1}(s/\Delta_{k,i})+\pi/2\right)+s\Delta_{k,i} \right],
\end{equation*}
which completes the proof.
\end{proof}

\vskip 0.2in

\bibliography{main}

\end{document}